\pdfoutput=1

\documentclass[11pt]{article}

\usepackage[final]{acl}
\usepackage{authblk}

\usepackage{times}
\usepackage{latexsym}
\usepackage{multirow}
\usepackage[T1]{fontenc}

\usepackage[utf8]{inputenc}

\usepackage{microtype}

\usepackage{inconsolata}
\usepackage{comment}

\usepackage{graphicx}
\usepackage{booktabs}
\usepackage{colortbl}
\usepackage{subcaption} 
\usepackage[utf8]{inputenc}
\usepackage{textcomp}
\DeclareUnicodeCharacter{2212}{\textminus}
\usepackage{amsmath}
\usepackage[inline]{enumitem}
\usepackage{pifont}
\newcommand{\cmark}{\ding{51}}
\newcommand{\xmark}{\ding{55}}
\usepackage{xcolor}
\usepackage{colortbl}
\definecolor{lightblue}{RGB}{230, 240, 255}
\definecolor{lightred}{RGB}{255, 230, 230}
\usepackage{threeparttable}
\usepackage{tcolorbox}
\usepackage{xcolor}

\newtcolorbox{findingbox}{
    colback=blue!5!white,
    colframe=blue!75!black,
    fonttitle=\bfseries,
}

\title{Does Finetuning with Scientific Data Increase Hallucinations? \\ A Multi-domain Factuality Evaluation of LLMs} 

\author{
  \textbf{Raia Abu Ahmad}\textsuperscript{1,2},
  \textbf{Nikolas Rauscher}\textsuperscript{1,2},
  \textbf{Ekaterina Borisova}\textsuperscript{1,2}, \\
  \textbf{Fabio Barth}\textsuperscript{1}, 
  \textbf{Georg Rehm}\textsuperscript{1,3},
  \textbf{Sebastian Möller}\textsuperscript{1,2}
}

\affil{
    \textsuperscript{1}Deutsches Forschungszentrum für Künstliche Intelligenz GmbH (DFKI), \\ 
    \textsuperscript{2}Technische Universität Berlin, 
    \textsuperscript{3}Humboldt-Universität zu Berlin \\ 
  \small{
   Corresponding author: \href{mailto:raia.abu_ahmad@dfki.de}{raia.abu\_ahmad@dfki.de}
  }
}

\begin{document}
\maketitle
\begin{abstract}
Large language models (LLMs) are increasingly used to communicate and explain scientific concepts, yet their tendency to hallucinate poses significant risks in this high stakes use-case. Prior scientific hallucination evaluation work remains largely restricted to the biomedical domain, treats hallucination as a binary task, and has not examined the growing family of scientifically fine-tuned LLMs. We address these gaps with \textsc{SciFactCheck}, a benchmark of 2,500 prompts across five scientific domains, paired with a modular evaluation framework targeting three factuality hallucination types: \emph{unverifiability}, \emph{overclaim}, and \emph{attribution}. Using a controlled minimal-pairing design, we evaluate 18 LLMs by comparing each scientifically fine-tuned model against its general-purpose base. Our results indicate that 1. \textbf{Scientifically fine-tuned models exhibit degraded factual reliability across all hallucination types and scientific domains}, and 2. \textbf{Fine-tuned models are internally less confident yet linguistically more assertive}. A human pilot study further reveals that current fact-checking tools show only modest agreement with expert judgments on scientific content, and that defining scientifically check-worthy claims remains contested even among human annotators. Our findings fundamentally challenge current methods of domain-specific fine-tuning for factuality and call for developing improved verification infrastructure for scientific content.
\end{abstract}

\section{Introduction}

\begin{figure*}[ht!]
\centering
\includegraphics[width=0.85\linewidth]{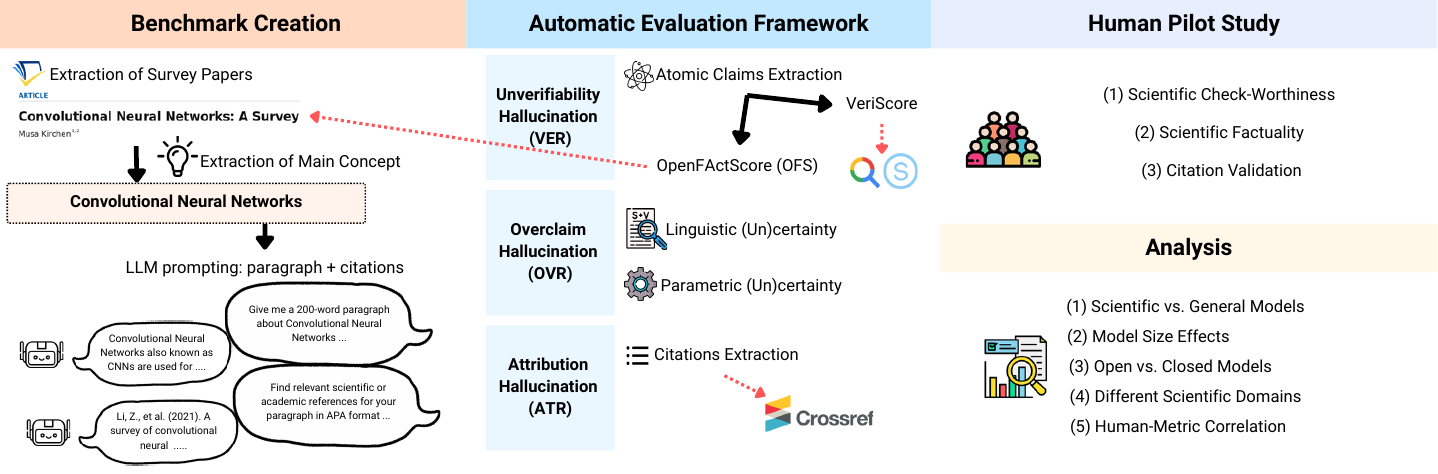}
    \caption{Illustration of the four-stage pipeline of our study.}
    \label{fig:pipeline}
\end{figure*}

Large language models (LLMs) have become central tools in scientific research and communication, with a growing ecosystem of models and agentic systems built specifically for scientific use cases~\cite{eger2025transforming,ghareeb_multi-agent_2026}. Students, researchers, and the general public increasingly rely on these systems to navigate complex or unfamiliar scientific concepts~\cite{alshomary2026llms}. However, LLMs are well-documented to generate factually incorrect content, a phenomenon broadly referred to as \emph{hallucinations}~\cite{maynez-etal-2020-faithfulness,augenstein2024factuality, hagendorff2024mapping, mishra2024finegrained}. Empirical work has consistently shown that long-tail knowledge is disproportionately prone to hallucinations~\cite{zong2024comparisonqa, zhao2024wildhallucinations}, and that longer open-ended responses exhibit more factual errors than shorter ones~\cite{zhao-etal-2025-response,vladika-etal-2025-facts}. These findings raise the stakes of deploying LLMs for text generation on scientific topics.

Despite this, existing work on open-ended LLM hallucination evaluation has been narrow in its scientific coverage, with most studies restricted to the biomedical domain~\cite{afzal-etal-2025-factbench, godbole2025verify} and none examining hallucinations in open-ended generation through the lens of \emph{error types}, a distinction that matters because different failure modes require different detection strategies and carry different consequences for users~\cite{mishra2024finegrained}. Equally important, no prior work has systematically evaluated the growing family of \emph{scientific LLMs}, which are fine-tuned on large collections of scientific publications to better serve domain experts. As these models become increasingly popular and widely deployed, understanding whether domain-specific fine-tuning actually improves factual reliability is a critical open question.

We address these gaps through the following contributions (see Figure~\ref{fig:pipeline}):
\begin{itemize}

\item \textsc{\textbf{SciFactCheck}}, a benchmark for open-ended, closed-book scientific text generation consisting of 2,500 prompts across five scientific domains\footnote{1. Engineering; 2. Life Sciences; 3. Physical Sciences and Mathematics; 4. Social and Behavioural Sciences; 5. Arts and Humanities.} grounded in peer-reviewed literature in English.\footnote{Our dataset is available at \url{https://huggingface.co/datasets/rabuahmad/scifactcheck}.}

\item \textbf{A modular evaluation framework} for scientific factuality hallucinations, targeting three hallucination types drawn from prior literature: \emph{unverifiability (\texttt{VER})}, \emph{overclaim (\texttt{OVR})}, and \emph{attribution (\texttt{ATR})}, each with dedicated automatic metrics based on existing libraries.\footnote{Our code for all steps of the study, including the evaluation framework, is available at \url{https://github.com/ryabhmd/SciFactCheck}.}

\item \textbf{A systematic evaluation of 18 LLMs} using a controlled minimal-pairing design, comparing scientific LLMs against the general-purpose foundation models they were built upon.

\item \textbf{A human pilot study} validating \texttt{VER} and \texttt{ATR} metrics against expert annotations across all scientific domains.
\end{itemize}

Based on the hallucination types, scientific domains, and used metrics, we find that scientifically fine-tuned models exhibit degraded factual reliability relative to general-purpose base models, challenging the premise that current domain-specific adaptation approaches improve factuality. We further find that scientifically fine-tuned models are internally less confident yet linguistically more assertive, a dissociation consistent with overclaiming behaviour. Our human study reveals an even deeper limitation: current automatic fact-checking tools show only modest agreement with expert judgments on scientific content, and annotators struggle to reach consensus on what constitutes a scientifically check-worthy claim, pointing to a fundamental gap in the field's verification infrastructure that goes beyond metric choice.

\section{Benchmark Construction}

\paragraph{Extracting Scientific Concepts.} We construct a dataset of scientific concepts across five high-level domains derived from the Open Research Knowledge Graph (ORKG) taxonomy\footnote{\url{https://orkg.org/fields}}: \begin{enumerate*}[label=\arabic*.] \item Engineering, \item Life Sciences, \item Physical Sciences and Mathematics, \item Social and Behavioural Sciences, and \item Arts and Humanities. \end{enumerate*} For the first four, we query the Semantic Scholar Open Research Corpus~\cite[S2ORC,][]{lo-etal-2020-s2orc}\footnote{S2ORC continuously updates its bulk dataset with new publications. We retrieved the data during October 2025.} for highly cited ($\geq$100 citations) survey and review papers, assuming that they contain consolidated, community-validated scientific knowledge. We extract concepts semi-automatically using Gemini-2.5-Flash~\cite{comanici2025gemini} on each title and abstract, followed by manual review to correct errors. For Arts and Humanities, where S2ORC returns insufficient results, we sample articles from the Internet Encyclopedia of Philosophy (IEP).\footnote{\url{https://iep.utm.edu/}} This process yields 500 scientific concepts per domain.\footnote{More details in Appendix~\ref{app:data-construction}.} Due to the nature of scientific knowledge, our design strategy does not use a fixed knowledge source and can be updated as science evolves. As an initial step, our benchmark's grounding in highly cited survey papers provides a stable, community-validated starting point from which more sophisticated verification pipelines can be built. 

\paragraph{Prompt Generation.} We use a two-step prompting strategy across all models and domains: first generating a paragraph about a scientific topic, then generating citations. 
For paragraph generation, we clarify our definition of factual scientific claims\footnote{\emph{Factual scientific claims are those that are considered true according to current consensus in the scientific community and/or according to historical facts about a scientific topic.}} and instruct the model to write a 200-word paragraph about the concept. We specify the length to reduce variance in the number of atomic claims and enable fairer metric comparisons. We use a zero-shot prompt, emulating the way a layperson would interact with an LLM~\cite{joseph-etal-2024-factpico}, without providing any external context. For citation generation, we follow the approach suggested by \citet{ravichander2025halogen} and instruct the model to provide references relevant to the paragraph it previously generated, formatted in APA style. We include two in-context examples to anchor the expected output format.\footnote{The full prompt templates are provided in Appendix~\ref{app:prompts}.}

\section{Scientific Factuality Hallucinations}
\label{sec:types}

We adopt the typology of \citet{pagnoni-etal-2021-understanding}, which distinguishes \emph{semantic frame errors} from \emph{content verifiability errors}, and survey prior work on general hallucination types~\cite{huang2024survey, li-etal-2024-dawn} and scientific error types~\cite{li-etal-2025-overview-scihal25, vendeville2025resource} to define six types of scientific factuality hallucinations (see Table~\ref{tab:hal-types} in Appendix~\ref{app:types}), selecting those recurring in at least two sources. We additionally include \texttt{ATR}, motivated by evidence of widespread citation fabrication in LLM outputs~\cite{walters2023fabrication, ravichander2025halogen, sakai2026hallucitation}. Since our evaluation targets open-ended closed-book generation, we focus exclusively on \emph{content verifiability errors} and assign dedicated automatic metrics to each type, prioritising those shown to correlate with human judgments~\cite{godbole2025verify}.

\subsection{Unverifiability Hallucination (\texttt{VER})}

\texttt{VER} refers to claims that are unsupported by trustworthy knowledge sources. In our case, these sources are peer-reviewed scholarly documents, which provide the best available descriptions of scientific consensus on a specific topic. Although scientific fact-checking (SFC) has specific challenges that don't apply to general fact-checking~\cite{tan2024faithful, deng2025next}, to the best of our knowledge, there is no open-source framework for evaluating factuality using scientific literature. Thus, we use two general-purpose metrics from different knowledge sources: FActScore~\cite{min-etal-2023-factscore} using its open-source implementation, OpenFActScore~\cite[OFS,][]{lage2025openfactscore}, and VeriScore~\cite{song-etal-2024-veriscore}. Both follow a retrieval-to-verification pipeline, where the output is deconstructed to atomic claims that are then queried against a knowledge source to retrieve relevant paragraphs for verification. We use VeriScore's claim extraction model for both metrics, as it has been shown to achieve higher agreement with human judgments~\cite{song-etal-2024-veriscore}.

For OFS, we construct a custom knowledge source from the full text of each survey paper. OFS reports the proportion of supported claims as the response-level score, only considering precision without accounting for recall. We report gamma scores ($\gamma=10$) to penalise short responses. For VeriScore, claims are verified against Google Search results via Serper API,\footnote{\url{https://serper.dev/}} restricted to documents published within the past year to avoid outdated evidence. Each claim is classified as \emph{supported} or \emph{unsupported} and an F1@K score is assigned to each output, rewarding factual precision (i.e., the fraction of supported claims) and sufficient claim coverage via recall. The latter is anchored to $K$, the median number of extracted claims across all model responses. 

\subsection{Overclaim Hallucination (\texttt{OVR})}

\texttt{OVR} errors overstate the scope, exclusivity, or certainty of a scientific claim. Definitions of \texttt{OVR} vary slightly in literature: \citet{li-etal-2024-dawn} describe it as statements that exceed what factual knowledge can support, while \citet{huang2024survey} as invalid claims due to subjective biases. Nonetheless, there is agreement that it is a distinct failure mode compared to \texttt{VER}: the latter constitutes scientifically check-worthy claims that cannot be verified against evidence sources, while the former is usually not check-worthy precisely because of its overclaiming nature. In scientific communication, overclaiming has been studied as a form of rhetorical manipulation, where the certainty or scope of findings is exaggerated relative to the underlying evidence \cite{li-etal-2017-nlp, patro-baruah-2021-simple}. This is closely related to the concept of \emph{model confidence} attribute of hallucinations~\cite{narayanan-venkit-etal-2024-audit}, which is the tendency of models to omit hedging language or uncertainty qualifiers. 

Given the absence of a dedicated metric for \texttt{OVR}, we focus on one of its aspects, \emph{(un)certainty}, and use two complementary proxies capturing different dimensions: \emph{linguistic certainty}, measuring how confidently a claim is expressed through lexical cues in the generated text, and \emph{parametric certainty}, measuring the model's internal confidence during generation. For the first, we use the certainty estimator of \citet{pei-jurgens-2021-measuring},\footnote{\url{https://pypi.org/project/certainty-estimator/}} which assigns each sentence a certainty score on a scale of 1 - 6, and outputs a response-level score consisting of the average across all sentences. For the second, we report length-normalised log probabilities of the output, following \citet{guerreiro-etal-2023-looking}, who show that sequence-level log probabilities correlate with hallucinated text. We compute the mean per-token log probability of each response, normalised by response length to avoid penalising longer outputs. 

\subsection{Attribution Hallucination (\texttt{ATR})}

\texttt{ATR} refers to fabricated or incorrect scientific attributions, commonly manifested as citations of non-existent or misrepresented publications. \texttt{ATR} concerns provenance rather than content, making it particularly damaging in scientific contexts where citations are the primary mechanism for grounding claims in the literature. Recent work has shown that LLMs hallucinate high percentages of generated citations~\cite{walters2023fabrication,ilter202617,sakai2026hallucitation}, with further studies showing the prevalence of this phenomenon~\cite{ravichander2025halogen,eger2025transforming,farhat2023trustworthy}.

Evaluating \texttt{ATR} in open-ended generation poses two challenges. First, to the best of our knowledge, no existing library or package supports citation hallucination evaluation in an open-ended generation setting at the time of writing. Second, models vary substantially in how many citations they generate per prompt. A model producing many citations may accumulate more valid ones simply by volume, making raw precision alone an insufficient basis for comparison. To address this, we introduce \textbf{CiteVal}, a metric inspired by VeriScore's F1@K that caps recall at a shared median target $M$ to enable fair cross-model comparison. This ensures all models are held to the same coverage expectation regardless of output length. Let $n$ be the total number of citations and $v \leq n$ the number of non-hallucinated citations per prompt:

\begin{equation}
\text{Precision} = \frac{v}{n}, \qquad \text{Recall}@M = \min\left(\frac{v}{M},\ 1\right)
\end{equation}

\noindent where $M$ is the median number of citations generated across all prompts and models. The primary metric is thus the rate of valid citations:

\begin{equation}
\text{CiteVal} = \text{F1}@M = \frac{2 \cdot \text{P} \cdot \text{R}@M}{\text{P} + \text{R}@M}
\end{equation}

computed per prompt and macro-averaged across all prompts to get a per-model score.

We verify citations against CrossRef~\cite{hendricks2020crossref} through a dual pipeline: after parsing APA-formatted citations with regular expressions, we use DOI-based and title-based lookup for validation. For DOI lookup, we normalise the DOI by removing whitespace and URL prefixes, and query CrossRef to verify its existence; if found, the retrieved title is compared to the citation using fuzzy matching\footnote{\url{https://github.com/seatgeek/thefuzz}}, and the citation is treated as non-hallucinated if similarity reaches a predefined threshold $\alpha = 0.9 $. For title lookup, we query the title directly, retrieve the top 10 candidates, and apply the same threshold to the best match.\footnote{We explain the choice of the $\alpha$ value in Appendix~\ref{app:citeval_threshold}.} A citation is classified as non-hallucinated if either strategy succeeds. Additional metadata (year, venue, authors) are not incorporated into our validation pipeline, since LLM-generated metadata frequently diverges from CrossRef records due to preprint and publication year discrepancies, venue abbreviations, and incomplete author lists.

\section{Experimental Setup}

We evaluate a set of LLMs spanning both general-purpose and scientifically fine-tuned model families in different parameter sizes (see Table~\ref{tab:models}). We identify four families of open-source models described as general scientific LLMs\footnote{More information on the training methods and corpora can be found in Appendix~\ref{app:sci-llms}.}: SciLitLLM~\cite{li2024scilitllmadaptllmsscientific}, fine-tuned on Qwen2.5~\cite{hui2024qwen2}; SciTulu~\cite{wadden-etal-2025-sciriff}, fine-tuned on Tulu-2~\cite{ivison2023camels}; S1-Base,\footnote{\url{https://huggingface.co/ScienceOne-AI/S1-Base-8B}, \url{https://huggingface.co/ScienceOne-AI/S1-Base-32B}} fine-tuned on Qwen3~\cite{yang2025qwen3}; and OpenScholar~\cite{asai2026synthesizing}, a retrieval-augmented (RAG) LM fine-tuned on Llama-3.1-8B\footnote{\url{https://huggingface.co/meta-llama/Llama-3.1-8B-Instruct}}. We further include two closed general-purpose models, GPT-4o-mini and Perplexity's Sonar, to provide reference points against proprietary systems. We prompt all models with a consistent generation configuration. Full implementation details are provided in Appendix~\ref{app:generation}.

\begin{table}[ht!]
\centering
\small
\resizebox{\columnwidth}{!}{
\begin{tabular}{llccc}
\toprule
\textbf{Model} & \textbf{Sci. Variant} & \textbf{Par. Sizes} & \textbf{Access} & \textbf{RAG} \\
\midrule
\texttt{Qwen2.5} & \texttt{SciLitLLM1.5} & 7B, 14B & Open & \xmark \\
\texttt{Llama2} \& \texttt{Tulu} & \texttt{SciTulu} & 7B, 70B & Open & \xmark \\
\texttt{Qwen3} & \texttt{S1-Base} & 8B, 32B & Open & \xmark \\
\texttt{Llama-3.1} & \texttt{OpenScholar} & 8B & Open & \cmark \\
\midrule
\texttt{GPT-4o-mini} & -- & -- & Closed & \xmark \\
\texttt{Sonar} & -- & -- & Closed & \cmark \\
\bottomrule
\end{tabular}}
\caption{LLMs evaluated in this study. Scientific models are paired with the general-purpose foundation model they were fine-tuned from for controlled comparison.}
\label{tab:models}
\end{table}
 
\section{Results}

\begin{table*}[ht!]
\small
\centering
\resizebox{0.9\textwidth}{!}{%
\begin{tabular}{l cc cc c}
\toprule
& \multicolumn{2}{c}{\texttt{VER}} & \multicolumn{2}{c}{\texttt{OVR}} & \multicolumn{1}{c}{\texttt{ATR}} \\
\cmidrule(lr){2-3} \cmidrule(lr){4-5} \cmidrule(lr){6-6}
\textbf{Model} & \textbf{OFS} & \textbf{VeriScore ($K=19$)} & \textbf{Ling.} & \textbf{Param.} & \textbf{CiteVal ($M=5, \alpha=0.9$)} \\
\midrule
\texttt{gpt-4o-mini} & 0.661  $\scriptstyle \pm .22$ & \cellcolor{lightblue} 0.728 $\scriptstyle \pm .23$ & 4.801 $\scriptstyle \pm .16$ & -- & 0.257 $\scriptstyle \pm .26$  \\
\texttt{perplexity-sonar} & \cellcolor{lightred} 0.580 $\scriptstyle \pm .22$ & 0.548 $\scriptstyle \pm .24$ & 4.901 $\scriptstyle \pm .08$ & -- & \cellcolor{lightblue} 0.451 $\scriptstyle \pm .30$\\
\midrule
\texttt{Llama-2-7b-chat}                  & 0.614 $\scriptstyle \pm .24$ & -- & \cellcolor{lightred} 4.747 $\scriptstyle \pm .17$ & \cellcolor{lightblue} -0.080 $\scriptstyle \pm .02$ & 0.173 $\scriptstyle \pm .24$ \\
\quad \texttt{tulu-2-7b}                       & 0.627 $\scriptstyle \pm .23$ & -- & 4.765 $\scriptstyle \pm .19$ & -0.166 $\scriptstyle \pm .05$ & 0.059 $\scriptstyle \pm .15$  \\
\quad \quad \texttt{scitulu-7b}                & 0.632 $\scriptstyle \pm .23$ & -- & 4.821 $\scriptstyle \pm .16$ & -0.228 $\scriptstyle \pm .05$ &\cellcolor{lightred}  0.013 $\scriptstyle \pm .07$  \\
\midrule
\texttt{Llama-2-70b-chat}                 & 0.622 $\scriptstyle \pm .21$ & -- & 4.776 $\scriptstyle \pm .16$ & -0.098 $\scriptstyle \pm .02$ & 0.227 $\scriptstyle \pm .27$  \\
\quad \texttt{tulu-2-70b}                & 0.656 $\scriptstyle \pm .21$ & 0.662 $\scriptstyle \pm .22$ & 4.793 $\scriptstyle \pm .17$ & -0.205 $\scriptstyle \pm .04$ & 0.300 $\scriptstyle \pm .28$ \\
\quad \quad \texttt{scitulu-70b}               & 0.655 $\scriptstyle \pm .22$ & 0.653 $\scriptstyle \pm .22$ & 4.802 $\scriptstyle \pm .18$ & -0.169 $\scriptstyle \pm .04$ & 0.177 $\scriptstyle \pm .21$ \\
\midrule
\texttt{Llama-3.1-8B-Instruct}           & 0.650 $\scriptstyle \pm .22$ & 0.604 $\scriptstyle \pm .23$ & 4.787 $\scriptstyle \pm .17$ & -0.360 $\scriptstyle \pm .06$ &  0.400 $\scriptstyle \pm .32$ \\
\quad \texttt{OpenScholar-8B}            & 0.628 $\scriptstyle \pm .23$ & 0.629 $\scriptstyle \pm .23$ & 4.793 $\scriptstyle \pm .17$ & \cellcolor{lightred} -0.374 $\scriptstyle \pm .06$ & 0.327 $\scriptstyle \pm .31$ \\
\midrule
\texttt{Qwen2.5-7B-Instruct}             & 0.652 $\scriptstyle \pm .22$ & 0.594 $\scriptstyle \pm .22$ & 4.782 $\scriptstyle \pm .19$ & -0.116 $\scriptstyle \pm .03$ & 0.250 $\scriptstyle \pm .28$ \\
\quad \texttt{SciLitLLM1.5-7B}           & 0.598 $\scriptstyle \pm .23$ & \cellcolor{lightred} 0.549 $\scriptstyle \pm .23$ & \cellcolor{lightred} 4.747 $\scriptstyle \pm .21$ & -0.173 $\scriptstyle \pm .05$ & 0.212 $\scriptstyle \pm .28$ \\
\midrule
\texttt{Qwen2.5-14B-Instruct}            & 0.667 $\scriptstyle \pm .21$ & -- & 4.807 $\scriptstyle \pm .16$ & -0.114 $\scriptstyle \pm .03$ & 0.323 $\scriptstyle \pm .31$ \\
\quad \texttt{SciLitLLM1.5-14B}          & 0.604 $\scriptstyle \pm .23$ & -- & 4.791 $\scriptstyle \pm .20$ & -0.105 $\scriptstyle \pm .04$ & 0.278 $\scriptstyle \pm .29$ \\
\midrule
\texttt{Qwen3-8B}                        & \cellcolor{lightblue} 0.671 $\scriptstyle \pm .21$ & -- & 4.850 $\scriptstyle \pm .14$ & -0.097 $\scriptstyle \pm .03$ & 0.274 $\scriptstyle \pm .26$  \\
\quad \texttt{S1-Base-8B}                & 0.605 $\scriptstyle \pm .22$ & -- & \cellcolor{lightblue} 4.912 $\scriptstyle \pm .10$ & -0.141 $\scriptstyle \pm .02$ & 0.164 $\scriptstyle \pm .20$ \\
\midrule
\texttt{Qwen3-32B}                       & 0.660 $\scriptstyle \pm .20$ & -- & 4.833 $\scriptstyle \pm .14$ & -0.149 $\scriptstyle \pm .03$ & 0.373 $\scriptstyle \pm .29$ \\
\quad \texttt{S1-Base-32B}               & 0.608 $\scriptstyle \pm .21$ & -- & 4.853 $\scriptstyle \pm .13$ & -0.108 $\scriptstyle \pm .02$ & 0.231 $\scriptstyle \pm .27$ \\
\bottomrule
\end{tabular}
}
\caption{Results of automatic evaluation metrics across models and factuality hallucination types. \textbf{VER} = Unverifiability (OFS and VeriScore; higher is better, i.e. more claims are verified), \textbf{OVR} = Overclaim (Linguistic and parametric uncertainty; higher is more certain), \textbf{ATR} = Attribution (CiteVal; higher is better, i.e. more citations are valid). Scientific LLMs are indented below their general-purpose base model. Logprobs are unavailable for closed-source models. VeriScore was evaluated on a subset of models due to computational constraints. \colorbox{lightblue}{Best} and \colorbox{lightred}{worst} scores per column are highlighted.}
\label{tab:results}
\end{table*}

Table~\ref{tab:results} presents results for all evaluated models across all hallucination types, aggregated across all five scientific domains. We report number of abstentions in responses, mean word per paragraph, and mean number of extracted claims and citations per model (Table~\ref{tab:response_stats}) and per domain (Table~\ref{tab:domain_stats}). The median number of extracted claims for VeriScore is $K=19$ and that of citations for CiteVal is $M=5$. Due to budget and time constraints, we compute VeriScore for 8 models, chosen based on the mean OFS scores for model pairs.\footnote{See more detailed model selection reasoning in Appendix~\ref{app:evaluation}.} We also report the most commonly used web domains for VeriScore evidence extraction in Appendix~\ref{app:evaluation}.

\paragraph{Overall Performance.} Closed models show varied performance: \texttt{GPT-4o-mini} achieves the highest VeriScore and \texttt{Sonar} achieves the highest CiteVal, consistent with its RAG architecture. However, \texttt{Sonar} also shows relatively low \texttt{VER} scores, with the lowest OFS out of all models. Among open models, \texttt{Qwen3-8B} achieves the highest OFS, \texttt{tulu-2-70b} the highest VeriScore, and \texttt{Llama-3.1-8B} the highest CiteVal. Overall, no specific model is consistently the best-performing across all metrics.

\begin{figure}[ht!]
\centering
\includegraphics[width=0.85\columnwidth]{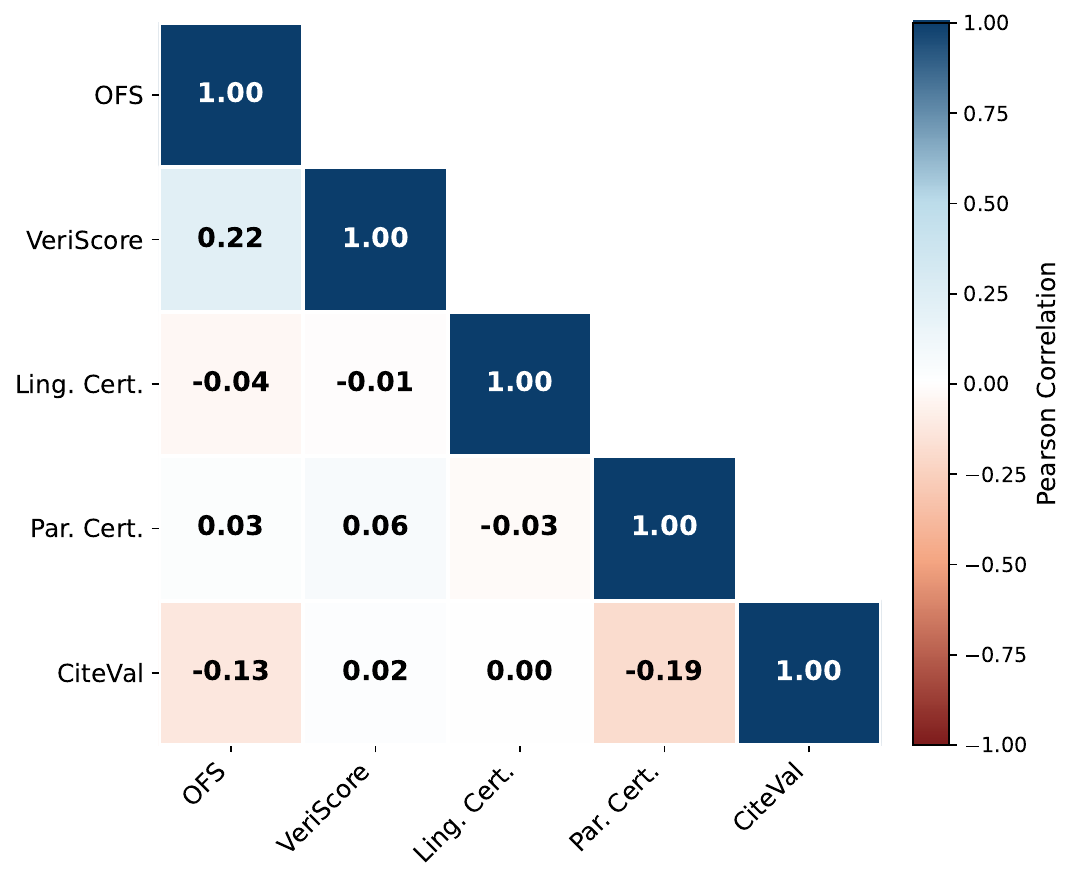}
    \caption{Pearson correlation matrix between metrics. Analysis includes six models with complete evaluations across 2,500 scientific concepts (15,000 observations). }
    \label{fig:correlations}
\end{figure}

\paragraph{Metric Correlations.} We compute pairwise Pearson correlations across all five metrics for the six models with complete evaluations (Figure~\ref{fig:correlations}). The results show that the three hallucination types are largely orthogonal: most pairwise correlations are around 0, indicating that metrics measure distinct errors. We note a divergence within \texttt{VER}, where OFS and VeriScore show a weak correlation ($r=0.22$), most probably attributed to the difference of evidence sources. We also note weak correlations between CiteVal and both OFS and par.~certainty, indicating that factual accuracy, citation behaviour, and model certainty show negative linear association. Linguistic certainty is not correlated with any other metric, suggesting that confident language is independent of factual accuracy.

\paragraph{Scientific vs. General Models.} To assess overall group differences, we apply the Mann-Whitney U test comparing general against scientific LLMs (Table~\ref{tab:overall_comparisons-small}). Results show a consistent pattern across metrics (all $p<0.001$): scientific LLMs underperform general LLMs on OFS, VeriScore, par.~certainty, and CiteVal, while scoring higher on ling.~certainty, with the most robust effect shown for par.~certainty (Cohen's $d=+0.36$). Per-pair comparisons using Wilcoxon signed-rank tests (Table~\ref{tab:paired_tests_appendix}) show a more nuanced picture: while CiteVal degradation seems to be consistent across all pairs, factual precision metrics vary by model family. For example, \texttt{OpenScholar} shows higher VeriScore than its base model, likely due to its RAG architecture. We also note that some scientific LLMs abstain more than their general counterparts.\footnote{See Appendix~\ref{app:evaluation}.}

\paragraph{Model Size Effect.} We assess overall size effects using Mann-Whitney U tests comparing small ($\leq8B$) against large ($>8B$) models (Table~\ref{tab:overall_comparisons-small}), finding that larger models are better in CiteVal, par.~certainty, and VeriScore, while OFS and ling.~certainty show no significant difference. Per-pair comparisons (Table~\ref{tab:size_effects_appendix}) reveal patterns similar to the scientific vs. general comparison: CiteVal shows consistent improvement in 6/7 pairs, par. certainty in 7/7 pairs, and OFS in 5/7 pairs (with mixed directionality). These results suggest model size effects exist but vary across model families and metrics.

\begin{table}[ht!]
\small
\centering
\resizebox{0.9\columnwidth}{!}{%
\begin{tabular}{l rr rl}
\toprule
\textbf{Metric} & \textbf{Group 1} & \textbf{Group 2}  & \textbf{p} & \textbf{d} \\
\midrule
\multicolumn{5}{l}{\textit{General vs. Scientific (N=9 gen, N=7 sci)}} \\
OFS & 0.667 & 0.632  & <0.001 & +0.15 \\
VeriScore & 0.647 & 0.600  & <0.001 & +0.13 \\
Ling.~Cert. & 4.837 & 4.869 & <0.001 & -0.15 \\
Par.~Cert. & -0.125 & -0.154 & <0.001 & +0.36 \\
CiteVal & 0.182 & 0.000 & <0.001 & +0.22 \\ 
\midrule
\multicolumn{5}{l}{\textit{Small vs. Large ($\leq$8B vs. $>$8B; N=9 small, N=7 large)}} \\
OFS & 0.652 & 0.667 & 0.036 & +0.04 \\
VeriScore & 0.611 & 0.686  & <0.001 & +0.28 \\
Ling.~Cert. & 4.846 & 4.851 & 0.059 & +0.04 \\
Par.~Cert. & -0.155 & -0.125  & <0.001 & +0.65 \\
CiteVal & 0.000 & 0.200 & <0.001 & +0.23 \\ 
\bottomrule
\end{tabular}
}
\caption{Overall group comparisons using Mann-Whitney U tests. Values are median scores. d = Cohen's d effect size. All tests use two-sided alternative.}
\label{tab:overall_comparisons-small}
\end{table}

\begin{figure*}[ht!]
\centering
\includegraphics[width=0.75\linewidth]{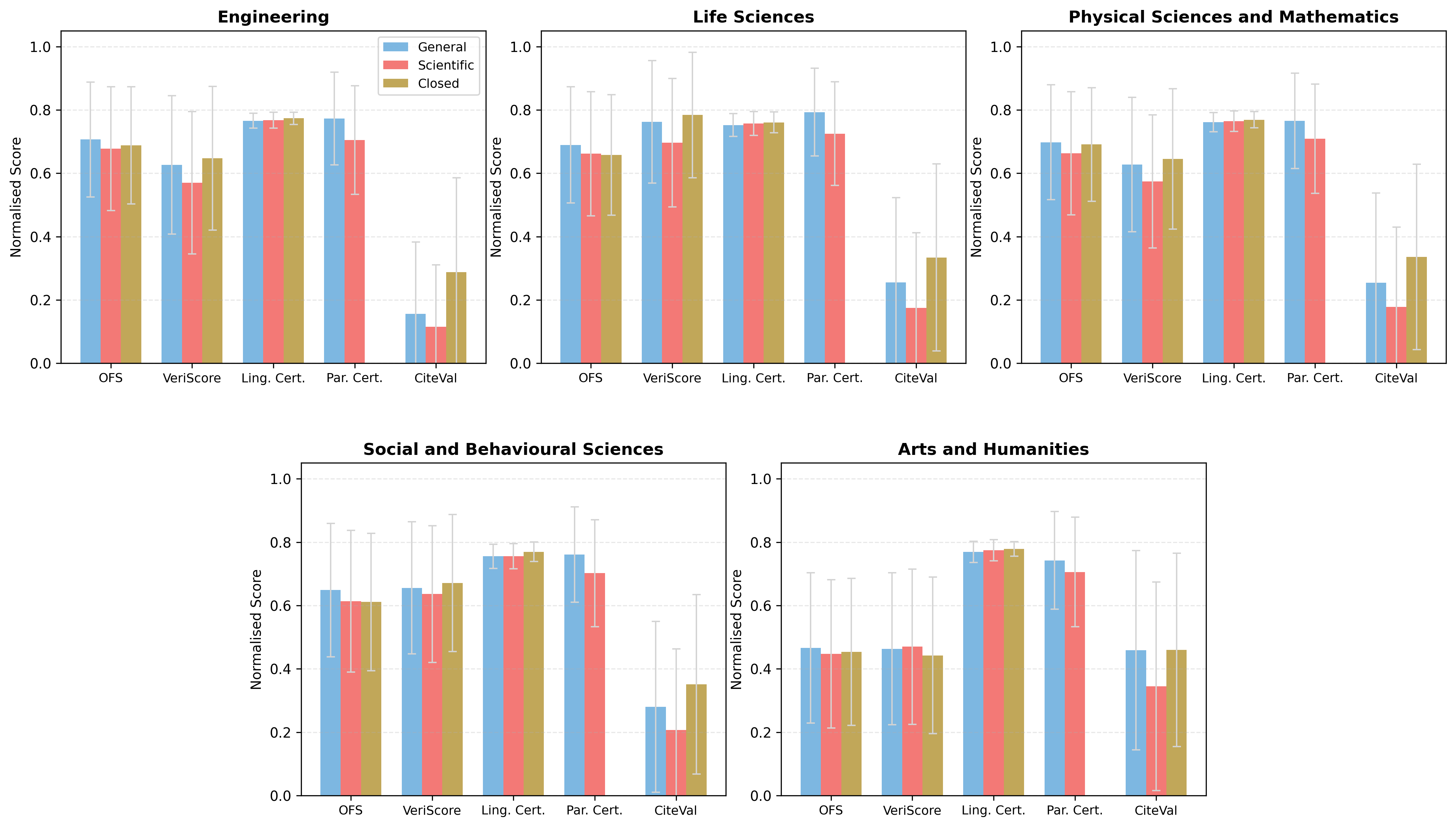}
    \caption{Performance across scientific domains by model type with normalised scores (0-1 scale). Error bars indicate standard deviation.
}
    \label{fig:domain_comparison}
\end{figure*}

\paragraph{Domain Variation.} Figure~\ref{fig:domain_comparison} shows performance across domains. We note that CiteVal scores are higher in Arts and Humanities, while ling.~and par.~certainty remain stable across all domains. OFS shows lower performance in Social and Behavioural Sciences and Arts and Humanities, and VeriScore follows a similar pattern, with Life Sciences having the highest scores and Arts and Humanities the lowest. However, we interpret \texttt{VER} differences with caution since the definition and verification of \emph{scientific facts} vary considerably across domains, and the availability and quality of evidence sources differ substantially between fields. As such, these epistemological differences limit direct comparisons of scores and we do not consider them indicative of any specific pattern.

\section{Human Validation Pilot Study}

We conduct a human validation study to assess the accuracy of our automatic metrics. Due to resource constraints, we restrict our study to \texttt{VER} and \texttt{ATR}.

\subsection{Validating \texttt{VER}}

We select two models representing the highest and lowest average performance: \texttt{GPT-4o-mini} and \texttt{Sonar}. We select 50 concepts (10 per domain) that cover both high- and low-scoring cases. To ensure annotators could evaluate claims correctly, we apply domain-specific topic filters.\footnote{Full concept list, scores, and topic filters are in Appendix~\ref{app:human_annotation}.} For each concept, we sample up to 10 claims per model from the same atomic claims used in the automatic evaluation, yielding 958 in total. Two annotators are recruited per domain with a minimal requirement of a bachelor's degree in the field. Annotators were asked to perform two tasks: (1) judge whether the claim is scientifically check-worthy; (2) for check-worthy claims, judge whether it is factual given a reference document comprising of the survey paper and web search snippets.\footnote{Annotation guidelines: \url{https://github.com/ryabhmd/SciFactCheck/blob/main/annotation_guidelines.pdf}} 

\paragraph{Inter-Annotator Agreement (IAA).} We report Cohen's $\kappa$ and raw agreement percentages due to label prevalence effects, as $\kappa$ underestimates agreement when one label dominates~\cite{artstein-poesio-2008-survey}. 
We use binary labels (\emph{check-worthy/not check-worthy}, \emph{supported/unsupported}) and see an overall fair agreement~\cite{landis1977application} for both check-worthiness ($\kappa$ = 0.29, 77\% raw) and factuality ($\kappa$ = 0.30, 72\% raw). IAA is highest for Physical Sciences and Mathematics on check-worthiness ($\kappa$ = 0.43) and for Life Sciences on factuality ($\kappa$ = 0.57), reflecting the greater clarity of verifiable claims in empirical domains. Unsurprisingly, Arts and Humanities shows the weakest check-worthiness ($\kappa$ = 0.17, 76\% raw) and factuality IAA ($\kappa$ = 0.06, 73\% raw), showcasing greater subjectivity in what constitutes a \emph{scientific fact}.\footnote{Detailed IAA scores and per-concept results are in Appendix~\ref{app:human_annotation}, where we also discuss examples of annotated claims and their scientific check-worthiness.}

\paragraph{Metric Validation.} We use the same metrics to compute agreement between \texttt{VER} metrics and human factuality judgments, where VeriScore shows higher results than OFS ($\kappa$ = 0.27, 69\% raw, vs. $\kappa$ = 0.05, 61\% raw). However, we interpret this with caution: annotators used VeriScore's web search snippets as the main reference, thus higher VeriScore-human agreement may reflect shared evidence rather than better metric quality. The divergence between OFS and VeriScore mirrors the weak correlation between them observed in Figure~\ref{fig:correlations}, reinforcing that evidence source selection fundamentally shapes evaluation outcomes. 

\subsection{Validating \texttt{ATR}}

We follow a similar process to validate CiteVal, adapted to citation-level judgments. We select two models, \texttt{SciTulu-7b} and \texttt{Llama-3.1-8b}, and for each domain select the two concepts with the highest and lowest average scores, yielding 10 concepts across the 5 domains and 115 extracted citations in total. For each citation, one annotator checked: (1) whether it was parsed correctly relative to the raw model output; (2) whether the cited title exists; (3) whether the cited DOI exists, and if so, whether it refers to the same title; and (4) whether the citation is topically related to its scientific concept.

Our results show that parsing is correct for 87\% of citations, with errors occurring primarily in venue extraction (15 citations), followed by title (11), authors (5), and DOI (2). At the adopted $\alpha=0.9$, CiteVal achieves an accuracy of 0.9 and an F1 score of 0.85 against manual validation (31 TP, 73 TN, 8 FP, 3 FN).\footnote{In Appendix~\ref{app:citeval_threshold}, we compare this with other $\alpha$ values.} Among hallucinated citations, the majority failed on title matching alone (74), while a small number failed due to a non-existent DOI (5) or a DOI that resolved to a different title than the one cited (1). Notably, a DOI was present in only 6 of the 115 citations overall, indicating that models rarely include DOIs in their generated citations and when they do, it's fabricated. Further inspection of the 35 citations verified as genuinely existing (TP + FN) shows that 25 were topically relevant to the concept they were cited for. Interestingly, all 10 non-relevant publications result from the Arts and Humanities domain.

\section{Discussion}

\paragraph{Scientifically fine-tuned models do not show improved factuality.} Despite being trained on scientific corpora, scientific LLMs did not outperform general-purpose models. While per-pair comparisons show some variation likely due to differences in fine-tuning 
strategies, this pattern holds at the aggregate level, challenging the premise that domain-specific training improves factuality. Several mechanisms are likely to contribute, such as catastrophic forgetting~\citep{luo2025empirical}, in which models lose previously acquired general knowledge during fine-tuning, as parameter updates overwrite existing factual representations. \citet{ghosal2024understanding} show that fine-tuning on new knowledge increases hallucination, particularly when new facts interact poorly with pretrained knowledge, and~\citet{gekhman-etal-2024-fine} similarly show that fine-tuning does not help models acquire new knowledge. Scientific fine-tuning may also suffer from data quality issues, as corpora can contain errors, outdated information, and unverified claims, and have been shown not to necessarily improve long-form factuality~\citep{newman2025curious}. Knowledge overshadowing~\citep{collins2026large,zhang-etal-2025-fever-law} could also be an explanation, where models struggle to reach scientific consensus because their training data contains extensive discussions of certain topics. 
More fundamentally, \citet{collins2026large} argue that LLMs lack the collective tacit knowledge humans acquire to function as trusted experts, suggesting that statistical learning is insufficient for genuine scientific \emph{understanding}. These results align with prior work showing that fine-tuning does not guarantee factual improvements~\citep{afzal-etal-2025-factbench,newman2025curious}, suggesting that current fine-tuning approaches require fundamental rethinking. Model size shows mixed associations with factuality, consistent with recent findings~\citep{zhao2024wildhallucinations, afzal-etal-2025-factbench}, cautioning against naïve scaling for mitigation.


\paragraph{Scientific models express confident language while being internally uncertain.} Overall, scientific LLMs exhibit higher ling.~certainty while showing lower par.~certainty. This shows a pattern of confident language masking internal uncertainty, likely because fine-tuning teaches models the stylistic conventions of scientific discourse rather than improving their knowledge: scientific papers typically employ assertive language, and models appear to learn this without acquiring the underlying epistemic rigour. Prior science communication work shows that misleading claims arise from increased linguistic confidence rather than incorrect content~\citep{li-etal-2017-nlp, patro-baruah-2021-simple}, suggesting that models may be replicating these patterns. The minor negative correlation between par.~and ling.~certainty further supports this, echoing concerns raised in prior hallucination research~\citep{narayanan-venkit-etal-2024-audit}. For scientific communication, where users may lack domain expertise to verify claims independently, this misalignment is particularly concerning.

\paragraph{Factuality is multidimensional and evidence source matters.} Pairwise correlations between metrics are low, suggesting that \texttt{VER}, \texttt{OVR}, and \texttt{ATR} represent independent dimensions of errors, and retrospectively justifying our multi-type framework. No single metric captures the full spectrum of failures, and optimising for one dimension does not necessarily improve another. Within \texttt{VER}, OFS and VeriScore showed weak correlation, highlighting that evidence source selection meaningfully shapes evaluation outcomes, which was further empirically grounded by our human pilot study. These results have practical implications for researchers evaluating scientific factuality of LLMs: metric choice is not neutral, and we recommend reporting multiple complementary metrics with different evidence sources rather than relying on a single one.

\paragraph{SFC remains an open challenge.} Our results reveal a broader limitation of current frameworks when applied to SFC. VeriScore, the stronger of the two metrics, achieved only modest agreement with human judgements, and human annotators themselves showed limited consensus on what constitutes a scientifically check-worthy claim. This points to a more fundamental gap: the scientific community lacks both a shared operational definition of scientific check-worthiness and the verification infrastructure to support it. Current pipelines like VeriScore retrieve evidence from the web, which in our data included sources of questionable authority such as social media  platforms (see Figure~\ref{fig:web-domains}), posing a significant concern when verifying scientific claims. We therefore argue that scientific factuality evaluation requires purpose-built verification tools grounded in curated scholarly sources, alongside explicit definitions of verifiability and check-worthiness for different scientific domains.

\section{Related Work}

\paragraph{Hallucination Detection.} Automatic methods usually adapt a \emph{decompose-and-verify} pipeline, which decomposes LLM generations into atomic claims and verifies each independently against external evidence, enabling claim-level judgments. FActScore~\cite{min-etal-2023-factscore} established this paradigm for open-ended biography generation, verifying atomic facts against Wikipedia; later extended by VeriScore~\cite{song-etal-2024-veriscore} to general long-form generation using web search evidence. OpenFactCheck~\cite{wang-etal-2025-openfactcheck} further unified several such pipelines into a single framework spanning claim extraction, evidence retrieval, and verification. This paradigm has since been extended in various directions, including converting claims and evidence to graph structures for verification~\cite{chen-etal-2025-graphcheck}, handling missing or incomplete facts~\cite{liu-etal-2025-verifact}, and bi-level verification frameworks using several evidence sources~\cite{zhao-etal-2025-response}.

The \emph{decompose-and-verify} paradigm is generally used because identifying specific failure modes is a prerequisite for developing targeted mitigation strategies~\cite{mishra2024finegrained,deemter-2024-pitfalls}. This is grounded in the broader distinction between \emph{faithfulness} (contradicting the input context) and \emph{factuality} (contradicting world knowledge) hallucinations~\cite{huang2024survey, li-etal-2024-dawn}. Faithfulness hallucinations have received more attention since the input context serves as a natural reference for verification, while factuality hallucinations are generally harder to detect~\cite{huang2024survey, li-etal-2024-dawn}. Following FActScore, most fine-grained factuality work targets open-ended biography generation using Wikipedia as an evidence source~\cite{min-etal-2023-factscore, zhao2024wildhallucinations, ravichander2025halogen}. However, previous work has consistently shown that long-tail entities, which are infrequent in training data or absent from Wikipedia, produce significantly more hallucinations. Such patterns are concerning for scientific content where many concepts are niche and domain-specific~\cite{afzal-etal-2025-factbench}, and LLMs have been shown to struggle with detecting and correcting scientific confabulations~\cite{wang-etal-2026-refact}. In this work, we explore this further in an open-ended closed-book generation setting.

\paragraph{SFC.} The predominant approach for SFC involves verifying claims extracted from real-world text against trusted evidence sources, mostly focusing on biomedicine and relying on abstracts for evidence~\cite{wadden-etal-2020-fact, wadden-etal-2022-scifact,sarrouti-etal-2021-evidence-based,saakyan-etal-2021-covid}. Recent work has broadened domain coverage to climate sciences~\cite{abu-ahmad-etal-2025-climatecheck} and natural language processing (NLP)~\cite{maab-yamagishi-2026-pushing}, the latter introducing full-text evidence and temporal verification dimensions. Although these benchmarks propose novel methods in SFC, none evaluate the factuality of claims generated by LLMs, which is the focus of our work.

\section{Conclusion}

We introduced \textsc{SciFactCheck}, a benchmark for evaluating scientific factuality hallucinations in open-ended LLM-generated text across five scientific domains, alongside a fine-grained evaluation framework targeting three hallucination types: \texttt{VER}, \texttt{OVR}, and \texttt{ATR}. Our systematic evaluation of 18 models revealed that, overall, scientific fine-tuning approaches do not show improved factual reliability relative to general-purpose base models across all hallucination types and domains, challenging the assumption of domain-specific adaptation for higher factuality. We further identified a dissociation between parametric and linguistic certainty in scientific models, suggesting that fine-tuning instils stylistic confidence without epistemic grounding. A human evaluation study indicated that automatic metrics of factuality show modest correlation with human judgments in the scientific domain, while also revealing that both automatic metrics and human annotators struggle with scientific check-worthiness, pointing to a fundamental gap in current SFC infrastructure.

\section*{Limitations}

\paragraph{Domain Classification.} Our benchmark relies on S2ORC's internal field classifier to assign papers to scientific domains. Despite filtering to papers assigned exclusively to a single domain cluster, the interdisciplinary nature of research cannot be fully eliminated. Some Engineering papers, e.g., concern applications that overlap with Computer Science. Domain-level comparisons should therefore be interpreted as approximate rather than strict.

\paragraph{Evidence Source Limitations.} Our \texttt{VER} evaluation relies on two knowledge sources: the full text of the survey paper (OFS) and web documents (VeriScore). Survey papers may not cover all relevant claims exhaustively, and VeriScore's web retrieval does not filter sources by citation count, publication venue, or web domain authority, and in practice retrieved documents from sources of limited scientific credibility (see Figure~\ref{fig:web-domains}). The limited availability of trustworthy evidence sources is a known open challenge in SFC~\cite{tan2024faithful}; prior work has shown that for the biomedical domain, Google Search results are comparable to PubMed-retrieved documents for verification quality~\cite{vladika-matthes-2024-comparing}, though whether this generalises across our five domains remains an open question, and our human evaluation is designed in part to assess these pipelines. Notably, our framework permits substituting evidence sources in future work, e.g., by expanding the OFS knowledge base to include papers cited by the source survey, or integrating purpose-built scientific verification pipelines such as that of \citet{tan-etal-2026-scitrue} as they become publicly available.

\paragraph{Overclaim Detection.} Overclaiming remains underexplored in the hallucination literature, and no dedicated automatic metric exists for detecting overclaims in open-ended generation. Our proxies, ling.~and par.~certainty, capture only one of its aspects according to previous definitions, and their validity as measures of overclaiming specifically has not been fully established. We leave \texttt{OVR} proxy validation as an important direction for future work.

\paragraph{CiteVal Limitations.} Our manual validation suggests that further experiments with the fuzzy-matching threshold and additional validation data could improve CiteVal's reliability. We additionally note that CiteVal verifies only whether a citation exists, not whether it supports the paragraph it accompanies. Extending it to check topical or evidential relevance is an important direction, though difficult in practice given that many publications lack public full-text access.

\paragraph{Training Data Contamination.} Our evaluation does not account for potential overlap between \textsc{SciFactCheck} concepts and model pretraining corpora. Probing this with existing tools~\cite{elazar2023s} could reveal whether well-represented concepts yield more factual outputs, consistent with frequency effects documented by \citet{zong2024comparisonqa}. This is particularly relevant for scientifically fine-tuned models, whose corpora are more likely to contain our source survey papers. However, many evaluated models do not disclose their full training data composition, and probing the subset that do would require significant additional effort. We leave this as a direction for future work.

\paragraph{Confounding Fine-tuning Factors.} Our study compares scientific LLMs against their general-purpose base models but does not control for differences in fine-tuning methodology, corpus curation, or data mixture ratios across the evaluated models. The observed performance degradation reflects the aggregate outcome of current scientific fine-tuning practices across diverse approaches rather than any single causal mechanism. Disentangling these factors would require further ablation studies and training multiple models under controlled conditions, which we leave for future work.

\paragraph{Human Study Scope.} Our \texttt{VER} human evaluation is limited to two models and 958 claims annotated by two annotators per domain, meaning per-domain metric agreement estimates should be interpreted with caution. Low IAA, particularly Arts and Humanities, reflects the genuine difficulty of scientific factuality judgement but limits the reliability of those results. Additionally, we require a minimum of a bachelor's degree rather than doctoral-level expertise, falling short of peer-review standards and potentially affecting judgement reliability for highly specialised concepts.

\section*{Ethical Considerations}

All survey papers for benchmark construction and knowledge base creation were collected in accordance with their open-access licensing status as determined by the Unpaywall schema~\cite{piwowar2018state}, restricting to Gold, Green, and Hybrid open-access papers. No paywalled or restricted content was used, ensuring compliance with copyright and licensing. All annotators were fully informed of the purpose and scope of the study and volunteered to assist without compensation. No personally identifiable information was collected from annotators, and participation was entirely voluntary with the option to withdraw at any time. Claude (Anthropic) was used to assist in polishing the manuscript and in writing portions of the code for model prompting and evaluation. All AI-assisted content was reviewed and verified by the authors.

\section*{Acknowledgments}
We thank the annotators who volunteered to help in the human pilot study: Mohammad Alsafadi, Judith Gilsbach, Tuğçe Soylu, Sepideh Baghaee, Markus Stricker, Muhammad Abu Ahmad, Max Upravitelev, and Marwa Nababteh. We also thank Vera Schmitt for consulting on the statistical tests and Eunhye Yun for discussions on catastrophic forgetting and other fine-tuning issues. Thank you to the anonymous reviewers for their valuable and constructive feedback. This work was supported by the consortium NFDI for Data Science and Artificial Intelligence (NFDI4DS)\footnote{\url{https://www.nfdi4datascience.de}} as part of the non-profit association National Research Data Infrastructure (NFDI e.\,V.). The consortium is funded by the Federal Republic of Germany and its states through the German Research Foundation (DFG) project NFDI4DS (no.~460234259). Further support was provided from the European Union’s Horizon Europe research and innovation programme under grant agreement No.~101189745 (HIVEMIND).


\bibliography{acl_latex}
\appendix

\section{Dataset and Prompt Details}
\label{sec:appendix}

\subsection{Further Details on Benchmark Construction}
\label{app:data-construction}

When filtering from S2ORC, we exclude papers without abstracts, and further filter to open-access publications with licenses permitting full-text reuse. Specifically, we keep those tagged as Gold, Green, or Hybrid according to the Unpaywall open-access status schema~\cite{piwowar2018state}. Papers in S2ORC are classified into 23 fine-grained scientific disciplines, which we query separately, mapping them to one of our five target domains manually. For each domain, we retain only publications assigned exclusively to disciplines within that domain's cluster, deliberately excluding multidisciplinary publications to ensure clean cross-domain comparability. We order results by descending publication year and prioritise recent ones, attempting to reduce data contamination in our experiments. The prompt used for scientific concept extraction is shown in Figure~\ref{box:gemini-prompt}. The results were manually reviewed by one author who discarded papers with at least one of the following exclusion criteria: \begin{enumerate}[label=\arabic*.] \item Repeated scientific concept; \item The paper is a subjective opinion piece about the concept rather than a literature review; \item The paper is not written in English; \item The paper is a follow-up of another review paper and thus does not contain the full context of information; \item The paper describes a questionnaire (i.e. \emph{survey} in that sense, which was prevalent in the Social Sciences domains).
\end{enumerate}
The IEP website is organised into five subsections: \begin{enumerate*}[label=\arabic*.] \item History of Philosophy, \item Metaphysics and Epistemology, \item Philosophical Traditions, \item Science, Logic, and Mathematics, and \item Value Theory. \end{enumerate*} We randomly select 100 articles from each subsection, yielding 500 philosophy concepts to represent the Arts and Humanities domain.


\begin{figure*}[ht!]
\center
\small
\begin{tcolorbox}[colback=gray!10, colframe=black, sharp corners=southwest,width=0.8\linewidth]

Given a scientific abstract of a survey paper about a scientific topic, your task is to extract the main scientific concept the abstract is about. Your response should consist of the main concept alone and nothing more. \\
If the given paper is not a survey paper about a given topic that can be used as a credible knowledge source about the scientific concept, you response must be None. \\
\\
Title of survey paper: \{title\}. \\
Abstract: \{abstract\}. \\
Main concept:
\end{tcolorbox}
\caption{Prompt used to extract the main scientific concept from the retrieved survey papers.}
\label{box:gemini-prompt}
\end{figure*}

\subsection{Prompt Templates}
\label{app:prompts}

Figure~\ref{box:prompt-paragraph} shows the prompt template used for the paragraph generation prompt and Figure~\ref{box:prompt-citation} shows the follow-up template used for the citation generation prompts. For paragraph generation, we follow \citet{afzal-etal-2025-factbench} in asking the model to respond based on its best knowledge, and \citet{song-etal-2024-veriscore}, in requesting that the output be fluent, coherent, and factual. We also specify the domain of the concept to disambiguate polysemous topics, for example, \emph{knowledge transfer} in education research vs. machine learning.

\begin{figure*}[ht!]
\center
\small
\begin{tcolorbox}[colback=gray!10, colframe=black, sharp corners=southwest,width=0.8\linewidth]

Scientific claims are those that are considered true according to current consensus in the scientific community and/or according to historical facts about a scientific topic. \\

Based on your best knowledge, give me a 200-word paragraph about the scientific concept \{scientific\_concept\} in the context of the \{field\} domain. \\

Content Instructions: \\
- The paragraph must be fluent, coherent, and consist only of factual scientific claims to the best of your knowledge. \\
- The paragraph must be generic and must NOT include names of specific studies, authors, or research groups. \\
- The paragraph must NOT include paper titles, journal or conference names, or DOIs. \\
- The paragraph must NOT include years linked to specific studies (e.g., ‘a 2015 study showed…’). \\
- Do NOT include any citations, reference markers, brackets, numbers, or footnotes inside the paragraph. \\
- Do NOT restate or explain these instructions. \\
- Do NOT repeat the same sentence, clause, or phrase multiple times. \\
\end{tcolorbox}
\caption{Paragraph generation prompt used for all models and domains.}
\label{box:prompt-paragraph}
\end{figure*}

\begin{figure*}[ht!]
\center
\small
\begin{tcolorbox}[colback=gray!10, colframe=black, sharp corners=southwest, width=0.8\linewidth]

Find relevant scientific or academic references in APA format that support the information presented in the following paragraph: \{paragraph\} \\ 

Use semicolons as separators, and list each APA reference without additional information. \\

Examples of APA formatted references: \\
Tang, F. \& Pierce J.W. (2014). Alzheimer's disease in young adults. Journal on Aging, 14(3), 220-243.; Iosua, S. L. (2017). Individual proximity, complex and coordinated collective movement. Journal of Comparative Psychology, 123(6), 126–134. https://doi.org/10.1167/com0000274 \\

** Note that complete citations MUST include author names, year of publication, title of publication, and publisher name**.

\end{tcolorbox}
\caption{Citation generation prompt used for all models and domains.}
\label{box:prompt-citation}
\end{figure*}

\section{Scientific Factuality Hallucination Types}
\label{app:types}

Table~\ref{tab:hal-types} displays six scientific factuality hallucination types according to prior research, each with a definition and an example illustrating it. In this paper, due to our focus on the open-ended text generation task, we focus on \emph{content verifiability error} types and leave the detection of \emph{semantic frame errors} to future work.

\begin{table*}[ht!]
\centering
\small
\resizebox{0.9\linewidth}{!}{
\begin{tabular}{lll}
\toprule
\textbf{Hallucination Type} & \textbf{Definition} & \textbf{Example}\\
\midrule
\multicolumn{3}{c}{\emph{Semantic Frame Errors}} \\
\midrule
Entity Error (\textcolor{red}{\texttt{ENT}})  & Error in named entity mentioned   & \emph{The theory of relativity was first} \\
\citet{huang2024survey,li-etal-2024-dawn, li-etal-2025-overview-scihal25} & in a claim. & \emph{published by \textcolor{red}{Newton}.}  \\
\citet{pagnoni-etal-2021-understanding} &  &   \\
\addlinespace

Relation Error (\textcolor{red}{\texttt{REL}}) & Error in a relation between  & \emph{Penicillin \textcolor{red}{was synthesized} by}     \\
\citet{huang2024survey,li-etal-2024-dawn} & two named entities in a claim. & \emph{Alexander Fleming.}   \\
\citet{pagnoni-etal-2021-understanding} & &  \\
\addlinespace

Numeric Error (\textcolor{red}{\texttt{NUM}})  & Error in a numerical subject  & \emph{Water boils at \textcolor{red}{90°C} at standard} \\
 \citet{li-etal-2025-overview-scihal25, pagnoni-etal-2021-understanding} & and/or object of a claim. & \emph{atmospheric pressure.}  \\ 
 \addlinespace
\midrule

\multicolumn{3}{c}{\emph{Content Verifiability Errors}} \\
\midrule
\addlinespace

Unverifiability (\textcolor{red}{\texttt{VER}}) & Claim cannot be verified against  & \emph{Mindfulness meditation activates}   \\
\citet{huang2024survey,li-etal-2024-dawn}  & a trusted knowledge source. & \emph{the parasympathetic nervous} \\
 \citet{li-etal-2025-overview-scihal25} & & \emph{system in 73\% of clinical cases.} \\
\addlinespace

Overclaim (\textcolor{red}{\texttt{OVR}}) & Exaggeration or overgeneralization   & \emph{Cognitive behavioural therapy}  \\ 
\citet{huang2024survey,li-etal-2024-dawn,li-etal-2025-overview-scihal25} & of a scientific claim in terms of  &  \emph{eliminates all symptoms of depression.} \\
\citet{vendeville2025resource} & scope, exclusivity, and/or certainty &  \\
\addlinespace

Attribution (\textcolor{red}{\texttt{ATR}}) & Fabricated scientific attribution & \emph{doi.org/10.1016/j.jhmt.2019.03.014} \\
 \citet{walters2023fabrication} &  &  \\
 \citet{ravichander2025halogen} &  &  \\
 \citet{sakai2026hallucitation,ilter202617} &  &  \\
\bottomrule
\end{tabular}}
\caption{Types of scientific hallucinations produced by LLMs based on previous research. In this study, we focus exclusively on content verifiability errors: \texttt{VER}, \texttt{OVR}, and \texttt{ATR}.}
\label{tab:hal-types}
\end{table*}

\section{Generation Configuration Details}
\label{app:generation}

\paragraph{Inference Frameworks.} Smaller open-source models are run using the HuggingFace Transformers library~\cite{wolf-etal-2020-transformers}. For larger open-source models, we use vLLM~\cite{kwon2023efficient} to enable efficient inference with tensor parallelization across multiple GPUs. Closed-source models are accessed via their respective provider APIs: GPT-4o-mini via the OpenAI API and Sonar via the Perplexity API. To unify API calls across providers, we use LiteLLM.\footnote{\url{https://github.com/BerriAI/litellm}} 

\paragraph{Hardware and Cost.} We prompted 16 local models using HuggingFace Transformers (v4.55.4) and vLLM (v0.11.0) across RTXA6000-SLT, RTXA6000, A100-80GB, H100, H100-Trail, and RTX3090 GPUs. Two closed-source models (GPT-4o-mini and Perplexity Sonar) were accessed via their respective APIs and are not included in the GPU hour count. Total prompting GPU compute for local models was approx.~530 GPU hours, and total cost for closed models was approx.~60 USD.

\paragraph{Hyperparameters.} We set the maximum number of new tokens to 1024 to avoid output truncation and apply each model's chat template where supported by its configuration. We enable sampling for all models, as recommended for generative instruction-following tasks requiring outputs longer than 20 tokens.\footnote{\url{https://huggingface.co/docs/transformers/main/en/generation_strategies}} We set the temperature to 0.3 across all models, following the same guidelines for tasks requiring factual rather than creative generation, striking a balance between determinism and output diversity. We apply a repetition penalty of 1.1 to reduce degenerate token repetitions.

\section{Scientific LLMs}
\label{app:sci-llms}

We report the training methods and used training corpora for the evaluated scientific LLMs in Table~\ref{tab:training-corpora}. We note that the most used datasets are: \begin{enumerate*}[label=\arabic*.] \item SciRIFF~\cite{wadden-etal-2025-sciriff}, which includes domains such as Material Sciences, Chemistry, Biomedicine, AI, Clinical, and Misc.; \item SciLitIns~\cite{li2024scilitllmadaptllmsscientific}, which is described as containing general science documents; and \item S2ORC, which is used for retrieval in \texttt{OpenScholar}. 
\end{enumerate*}

\begin{table*}[ht!]
\centering
\small
\begin{tabular}{llll}
\toprule
\textbf{Model} & \textbf{Stage} & \textbf{Dataset} & \textbf{Size}  \\
\midrule
\multirow{6}{*}{\texttt{SciLitLLM}}
    & CPT & In-house textbooks & 73k / 10B tok  \\
    & CPT & In-house journals  & 625k / 2.7B tok  \\
    & CPT & RedPajama          & 11B tok  \\
    & SFT & SciLitIns          & 93k inst \\
    & SFT & SciRIFF            & 70k inst  \\
    & SFT & Infinity-Instruct  & 3M inst  \\
\midrule
\multirow{2}{*}{\texttt{SciTulu}}
    & SFT & SciRIFF-train-mix  & 70k inst  \\
    & SFT & Tulu v2 SFT mixture & —  \\
\midrule
\multirow{2}{*}{\texttt{S1-Base}}
    & CPT & Scientific papers  & 170M docs \\
    &     &                    &            \\
    & SFT & Scientific reasoning instances & —  \\
\midrule
\multirow{3}{*}{\texttt{OpenScholar}}
    & Retrieval & peS2o v3 / S2ORC & 45M papers  \\
    & SFT & Synthetic pipeline data & 130k inst  \\
    & SFT & SciRIFF  & —  \\
\bottomrule
\end{tabular}
\caption{Disclosed training corpora for scientifically fine-tuned models evaluated in this work. CPT = continual pre-training; SFT = supervised fine-tuning.}
\label{tab:training-corpora}
\end{table*}

\section{Metric Implementation Details}
\label{app:metric-implementation}

\paragraph{Custom Knowledge Sources for OFS.} We process each PDF of the original survey articles using Docling v.2.61.2~\cite{livathinos2025docling} and MinerU v1.3.2~\cite{wang2024mineru}, extracting the text as markdown and converting it to JSONL to match the required input format, where each paragraph in the full text is treated as a retrievable unit. For IEP articles, we use the BeautifulSoup package v4.13.3\footnote{\url{https://beautiful-soup-4.readthedocs.io/en/latest/}} to scrape each article and similary convert it to JSONL formatting for retrieval. We verify claims by first retrieving the top 5 paragraphs per claim via GTR-T5\footnote{\url{https://huggingface.co/sentence-transformers/gtr-t5-large}} and validating with Gemma-7B-Instruct\footnote{\url{https://huggingface.co/google/gemma-7b-it}} using chain-of-thought reasoning.

\paragraph{Hardware and Cost.} OFS evaluation was run using Gemma-7B-it as the atomic fact verification model across A100-80GB and H100-Trail GPUs, requiring approx.~298 GPU hours in total. VeriScore evaluation required an approx.~180 GPU hours for claim extraction (Mistral-based) and claim verification (Llama-3-based) steps across the 18 evaluated models; evidence retrieval was API-bound and required no GPU compute. For \texttt{OVR} and \texttt{ATR} evaluation, only CPUs were used. Total evaluation cost was approx.~478 GPU hours, as well as approx.~400 USD for Serper API usage.

\section{Model Responses and Additional Evaluation Results}
\label{app:evaluation}

We report response statistics per model (Table~\ref{tab:response_stats}) and domain (Table~\ref{tab:domain_stats}), including number of abstentions for both the paragraph and citation generation prompts, mean number of words per response to the paragraph generation task, mean number of extracted claims and mean number of extracted citations. In reporting abstentions, we include instances where no atomic claims were extracted and no citations were extracted. In citation generation, some models, such as \texttt{Tulu-2-7b} and \texttt{SciTulu-7b}, tend to output the two citations given in our two-shot prompt, which we don't count as a proper response, since it is a mere replication of the input samples. We count those as instances with no extracted citations, but not as abstentions. 

We show average OFS results of each generic-scientific model pair in Table~\ref{tab:pair_rankings}, on which base our model selection for running VeriScore. To limit our bias in interpreting VeriScore results, we choose the top and bottom pairs and add the \texttt{Llama-3.1-8B-Instruct}-\texttt{OpenScholar-8B} pair to maximise model family diversity. 

\begin{figure*}[ht!]
\centering
\includegraphics[width=0.95\textwidth]{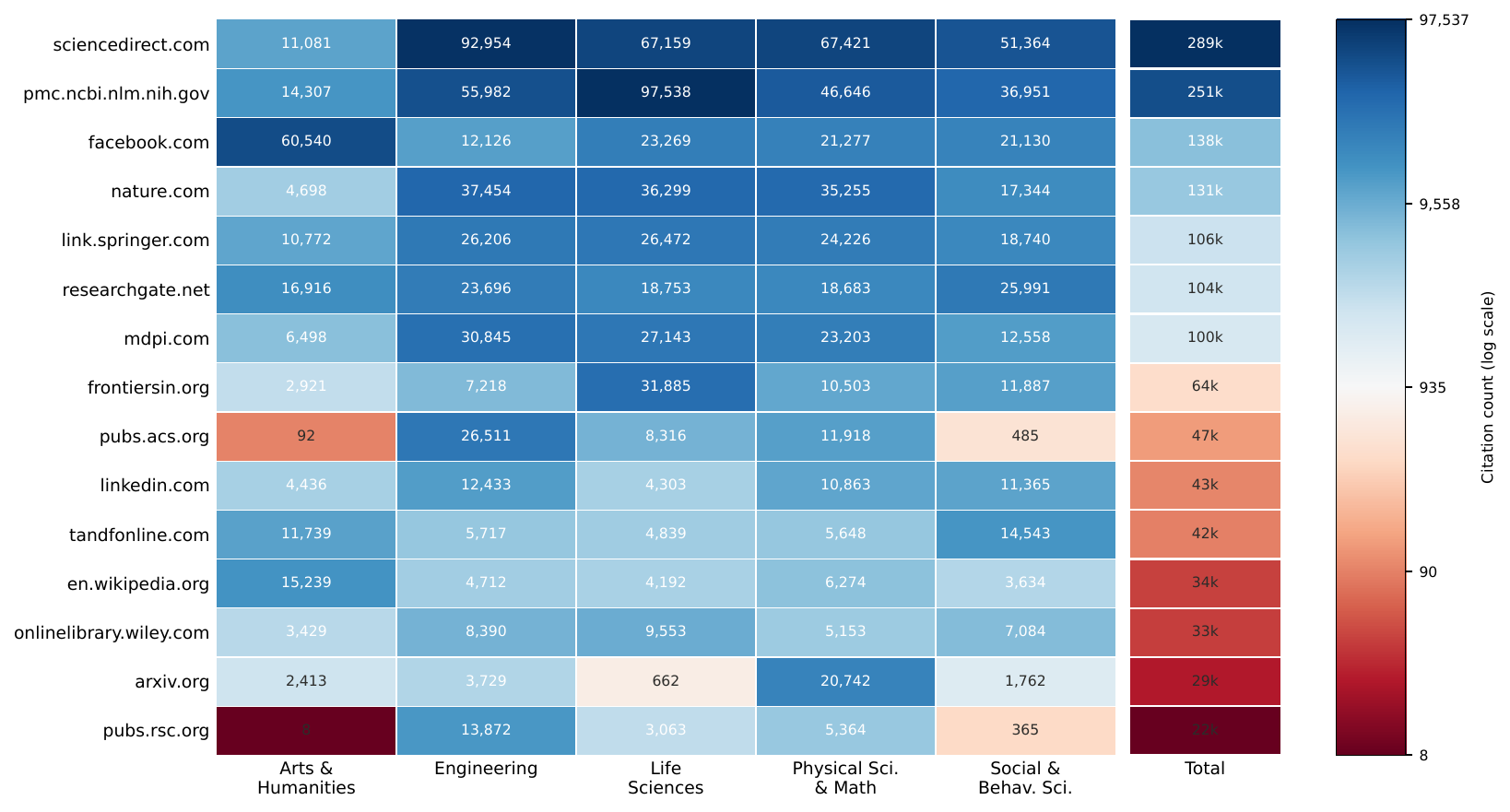}
    \caption{Top 15 web domains used by VeriScore via Serper API to fact-check the given atomic claims per domain, aggregated count across all evaluated model outputs.}
    \label{fig:web-domains}
\end{figure*}

Figure~\ref{fig:web-domains} shows the most frequent web domains used by VeriScore to verify atomic claims extracted from the paragraph generation responses. We note that, for the most part, scientific websites are used to extract evidence (e.g. \url{https://www.sciencedirect.com} and \url{https://www.nature.com}). However, social media websites are also used, especially for scientific domains where not enough public peer-reviewed sources exist, such as Arts and Humanities, where the top website used to extract evidence is \url{https://www.facebook.com}. 

\begin{table*}[ht!]
\centering
\small
\resizebox{0.9\textwidth}{!}{%
\begin{tabular}{lccccc}
\toprule
\textbf{Model} & \textbf{Abst. (paragraphs)} & \textbf{Words (paragraphs)} & \textbf{Claims} & \textbf{Abst. (citations)}  & \textbf{Citations} \\
&  & \textbf{(mean)} & \textbf{(mean ± SD)} & & \textbf{(mean ± SD)} \\
\midrule
\texttt{GPT-4o-mini} & 0 & 188.8 & 20.5 ± 5.9 & 1 & 5.5 ± 1.0 \\
\texttt{Sonar} & 3 & 184.9 & 25.8 ± 8.3 & 226 & 4.0 ± 2.2 \\
\midrule
\texttt{Llama-3.1-8B-Instruct} & 2 & 147.9 & 17.8 ± 4.7 & 5 & 9.3 ± 5.2 \\
\quad \texttt{OpenScholar-8B} & 0 & 155.7 & 19.2 ± 5.5 & 21 & 5.7 ± 6.2 \\
\midrule
\texttt{Qwen2.5-14B-Instruct} & 1 & 150.6 & 18.5 ± 4.8 & 1 & 5.7 ± 1.7 \\
\quad \texttt{SciLitLLM1.5-14B} & 4 & 115.7 & 15.4 ± 5.8 & 40 & 5.5 ± 3.7 \\
\midrule
\texttt{Qwen2.5-7B-Instruct} & 0 & 135.6 & 17.1 ± 4.6 & 1 & 8.0 ± 4.1 \\
\quad \texttt{SciLitLLM1.5-7B} & 1 & 112.1 & 14.4 ± 5.6 & 230 & 5.6 ± 8.0 \\
\midrule
\texttt{Qwen3-32B} & 0 & 179.7 & 24.7 ± 5.7 & 3 & 7.7 ± 2.4 \\
\quad \texttt{S1-Base-32B} & 6 & 204.4 & 29.6 ± 7.5 & 304 & 5.7 ± 3.4 \\
\midrule
\texttt{Qwen3-8B} & 1 & 180.2 & 22.2 ± 5.9 & 3 & 12.5 ± 7.3 \\
\quad \texttt{S1-Base-8B} & 1 & 183.4 & 24.6 ± 7.6 & 110 & 5.9 ± 4.1 \\
\midrule
\texttt{Llama-2-70b-chat} & 0 & 263.7 & 29.3 ± 8.8 & 0 & 8.6 ± 2.7 \\
\quad \texttt{Tulu-2-70b} & 0 & 164.3 & 17.6 ± 5.3 & 7 & 5.1 ± 3.5 \\
\quad \quad \texttt{SciTulu-70b} & 0 & 159.8 & 17.3 ± 5.2 & 7 & 2.1 ± 2.6 \\
\midrule
\texttt{Llama-2-7b-chat} & 0 & 158.7 & 17.4 ± 5.4 & 0 & 6.6 ± 2.7 \\
\quad \texttt{Tulu-2-7b} & 1 & 146.7 & 16.4 ± 6.1 & 1223 & 1.1 ± 1.9 \\
\quad \quad \texttt{SciTulu-7b} & 0 & 160.4 & 16.9 ± 6.1 & 91 & 0.2 ± 1.0 \\
\bottomrule
\end{tabular}
}
\caption{Response characteristics per model. Abst. = abstention number on the paragraph/citation generation prompt (out of 2,500 prompts per model), includes instances where no atomic claims or citations were extracted; Claims and Citations show the average extractions per prompt across all 2,500 concepts. Scientific models are grouped below their general-purpose base model.}
\label{tab:response_stats}
\end{table*}

\begin{table*}[ht!]
\centering
\small
\resizebox{0.9\textwidth}{!}{%
\begin{tabular}{lccccc}
\toprule
\textbf{Domain} & \textbf{Abst. (paragraphs)} & \textbf{Words (paragraphs)} & \textbf{Claims} & \textbf{Abst. (citations)} & \textbf{Citations} \\
&  & \textbf{(mean)} & \textbf{(mean ± SD)} & & \textbf{(mean ± SD)} \\
\midrule
Engineering & 4 & 165.2 & 22.3 ± 7.7 & 412 & 5.2 ± 3.8 \\
Life Sciences & 1 & 161.7 & 21.9 ± 7.8 & 378 & 5.5 ± 4.0 \\
Physical Sciences \& Math & 0 & 168.7 & 20.7 ± 7.5 & 485 & 5.4 ± 4.0 \\
Social \& Behavioural Sci. & 8 & 169.1 & 19.6 ± 7.2 & 355 & 6.4 ± 5.0 \\
Arts and Humanities & 7 & 166.5 & 16.9 ± 6.8 & 643 & 6.6 ± 7.0 \\
\bottomrule
\end{tabular}
}
\caption{Response characteristics per domain, averaged across all models and 500 concepts per domain, equaling 9000 prompt outputs per domain. Abst. = abstention number on the paragraph/citation generation prompt, includes instances where no atomic claims or citations were extracted.}
\label{tab:domain_stats}
\end{table*}

\begin{table}[h]
\small
\centering
\begin{tabular}{clc}
\hline
\textbf{Rank} & \textbf{Pair} & \textbf{Avg OFS}  \\
\hline
1* & \texttt{tulu-2-70b},  & 0.6551 \\
 &  \texttt{scitulu-70b} &   \\
 
2* & \texttt{Llama-3.1-8B-Instruct}, & 0.6388 \\
 & \texttt{Llama-3.1\_OpenScholar-8B} &   \\
 
3 & \texttt{Qwen3-8B} & 0.6379  \\
 & \texttt{S1-Base-8B} &   \\

4 & \texttt{Qwen2.5-14B-Instruct} & 0.6348  \\
 & \texttt{SciLitLLM1.5-14B} &  \\

5 & \texttt{Qwen3-32B},  & 0.6333 \\
 & \texttt{S1-Base-32B} &  \\

6 & \texttt{tulu-2-7b}, & 0.6292 \\
 & \texttt{scitulu-7b} &   \\

7* & \texttt{Qwen2.5-7B-Instruct},  & 0.6251 \\
 & \texttt{SciLitLLM1.5-7B} &   \\

\hline
\end{tabular}
\caption{General-scientific pair rankings by average OFS results for open-source models. Selected pairs for VeriScore evaluation are marked with *.}
\label{tab:pair_rankings}
\end{table}

\section{Detailed Statistical Test Results}

We conducted all tests using scipy v1.15.3\footnote{\url{https://scipy.org/}} and scikit-learn v1.2.0\footnote{\url{https://scikit-learn.org/stable/}}. Table~\ref{tab:overall_comparisons} presents the group comparison results using the Mann-Whitney U tests in more detail, and Table~\ref{tab:paired_tests_appendix} shows detailed results of the paired Wilcoxon signed-rank tests between the seven general vs. scientific pairs on all metrics. Table~\ref{tab:size_effects_appendix} presents results of the same test for pairs of different model sizes from the same family. We choose this test as well as the Mann-Whitney U test following recommendations for statistical testing in NLP evaluation settings where metric distributions are non-normal~\cite{dror-etal-2018-hitchhikers, peyrard-etal-2021-better}. Factuality scores such as OFS gamma scores and VeriScore F1@K tend to cluster at the extremes or follow skewed distributions, violating the normality assumption required by parametric tests such t-tests or ANOVA. We therefore rely on non-parametric tests throughout our analysis. For overall group comparisons, where we compare all general-purpose models against all scientifically fine-tuned models, or all small models against all large models, observations are unpaired since we are comparing groups rather than matched pairs. We use the \emph{Mann-Whitney U test} (also known as the \emph{Wilcoxon rank-sum test}) for these comparisons. For paired comparisons, where we compare a specific scientifically fine-tuned model directly against the general-purpose foundation model it was built upon, observations are paired at the prompt level. We use the \emph{Wilcoxon signed-rank test} for these comparisons, which accounts for the paired structure of the data. When we compare more than two models simultaneously (e.g., across size groups), we apply the \emph{Friedman test} followed by Wilcoxon signed-rank tests with Bonferroni correction for post-hoc pairwise comparisons. We also report Cohen's d as a measure of effect size alongside p-values, as statistical significance alone can be misleading given the large sample sizes ($N=2,500$ prompts per model)~\cite{peyrard-etal-2021-better}.

\begin{table}[ht!]
\small
\centering
\resizebox{\columnwidth}{!}{%
\begin{tabular}{l rr rrl}
\toprule
\textbf{Metric} & \textbf{Group 1} & \textbf{Group 2} & \textbf{U} & \textbf{p} & \textbf{d} \\
\midrule
\multicolumn{6}{l}{\textit{General vs. Scientific (N=9 gen, N=7 sci)}} \\
OFS & 0.667 & 0.632 & 183,667,700.5 & <0.001 & +0.15 \\
VeriScore & 0.647 & 0.600 & 20,178,132 & <0.001 & +0.13 \\
Ling. Cert. & 4.837 & 4.869  & 147,495,607 & <0.001 & -0.15 \\
Par. Cert. & -0.125 & -0.154  & 208,606,795 & <0.001 & +0.36 \\
CiteVal & 0.182 & 0.000 & 190,622,048.5 & <0.001 & +0.22 \\ 
\midrule
\multicolumn{6}{l}{\textit{Small vs. Large ($\leq$8B vs. $>$8B; N=9 small, N=7 large)}} \\
OFS & 0.652 & 0.667 & 194,302,440.5 & 0.036 & +0.04 \\
VeriScore & 0.611 & 0.686 & 20,939,776.5 & <0.001 & +0.28 \\
Ling. Cert. & 4.846 & 4.851 & 194,700,144.5 & 0.059 & +0.04 \\
Par. Cert. & -0.155 & -0.125  & 143,451,372 & <0.001 & +0.65 \\
CiteVal & 0.000 & 0.200  & 167,675,674 & <0.001 & +0.23 \\ 
\bottomrule
\end{tabular}
}
\caption{Overall group comparisons using Mann-Whitney U tests. Values are median scores. U = Mann-Whitney U statistic. d = Cohen's d effect size. All tests use two-sided alternative.}
\label{tab:overall_comparisons}
\end{table}

\begin{table*}[ht]
\small
\centering
\resizebox{0.9\textwidth}{!}{%
\begin{tabular}{ll l rr rrr l}
\toprule
\textbf{General Model} & \textbf{Scientific Model} & \textbf{Metric} & \textbf{Gen Med} & \textbf{Sci Med} & \textbf{W} & \textbf{p-value} & \textbf{d} & \textbf{Sig} \\
\midrule
\texttt{Llama-3.1-8B} & \texttt{OpenScholar-8B} & OFS & 0.683 & 0.650 & 1,272,651.5 & <0.001 & +0.09 & *** \\
 &  & VeriScore & 0.619 & 0.651 & 1,290,416.5 & <0.001 & -0.11 & *** \\
 &  & Ling. Cert. & 4.821 & 4.820 & 1,523,463 & 0.272 & -0.03 & ns \\
 &  & Par. Cert. & -0.358 & -0.370 & 1,165,675 & <0.001 & +0.25 & *** \\
 &  & CiteVal & 0.364 & 0.250 & 797,696 & <0.001 & +0.24 & *** \\
\midrule
\texttt{Qwen2.5-14B} & \texttt{SciLitLLM1.5-14B} & OFS & 0.690 & 0.615 & 991,452.5 & <0.001 & +0.28 & *** \\
 &  & Ling. Cert. & 4.860 & 4.845 & 1,456,492.5 & 0.003 & +0.09 & ** \\ 
 &  & Par. Cert. & -0.112 & -0.097 & 1,156,032.0 & <0.001 & -0.28 & *** \\
 &  & CiteVal & 0.200 & 0.200 & 745,916.0 & <0.001 & +0.15 & *** \\
\midrule
\texttt{Qwen2.5-7B} & \texttt{SciLitLLM1.5-7B} & OFS & 0.684 & 0.600 & 1,053,141.5 & <0.001 & +0.24 & *** \\
 &  & VeriScore & 0.611 & 0.556 & 1,126,295 & <0.001 & +0.20 & *** \\
 &  & Ling. Cert. & 4.845 & 4.806 & 1,276,050 & <0.001 & +0.18 & *** \\
 &  & Par. Cert. & -0.114 & -0.168 & 163,798 & <0.001 & +1.50 & *** \\
 &  & CiteVal & 0.167 & 0.0 & 668,373 & <0.001 & +0.14 & *** \\
\midrule
\texttt{Qwen3-32B} & \texttt{S1-Base-32B} & OFS & 0.680 & 0.625 & 925,694 & <0.001 & +0.26 & *** \\
 &  & Ling. Cert. & 4.877 & 4.898 & 1,260,786.5 & <0.001 & -0.15 & *** \\
 &  & Par. Cert. & -0.147 & -0.107 & 160,347 & <0.001 & -1.44 & *** \\
 &  & CiteVal & 0.333 & 0.167 & 492,108.5 & <0.001 & +0.51 & *** \\
\midrule
Qwen3-8B & S1-Base-8B & OFS & 0.706 & 0.630 & 866,699.5 & <0.001 & +0.30 & *** \\
 &  & Ling. Cert. & 4.892 & 4.940 & 704,687 & <0.001 & -0.53 & *** \\
 &  & Par. Cert. & -0.095 & -0.141 & 77,100 & <0.001 & +1.80 & *** \\
 &  & CiteVal & 0.200 & 0.118 & 487,172 & <0.001 & +0.47 & *** \\
\midrule
\texttt{tulu-2-70b} & \texttt{scitulu-70b} & OFS & 0.682 & 0.684 & 1,439,862 & 0.926 & +0.00 & ns \\
 &  & VeriScore & 0.688 & 0.688 & 1,405,490 & 0.005 & +0.04 & * \\
 &  & Ling. Cert. & 4.826 & 4.839 & 1,429,897 & <0.001 & -0.05 & *** \\ 
 &  & Par. Cert. & -0.202 & -0.166 & 391,958 & <0.001 & -0.89 & *** \\
 &  & CiteVal & 0.333 & 0.0 & 290,710.5 & <0.001 & +0.49 & *** \\
\midrule
\texttt{tulu-2-7b} & \texttt{scitulu-7b} & OFS & 0.647 & 0.651  & 1,416,246.5 & 0.010 & -0.02 & ns \\
 &  & Ling. Cert. & 4.812 & 4.854 & 1,035,821 & <0.001 & -0.31 & *** \\
 &  & Par. Cert. & -0.162 & -0.226 & 184,745 & <0.001 & +1.35 & *** \\
 &  & CiteVal & 0.0 & 0.0 & 10,115 & <0.001 & +0.39 & *** \\
\bottomrule
\end{tabular}
}
\caption{Paired Wilcoxon signed-rank tests comparing general models to their scientifically fine-tuned counterparts. Each pair evaluated on 2,500 scientific concepts. Positive differences indicate general model outperforms scientific model. \textbf{Gen Med/Sci Med}: Median scores for general and scientific models. \textbf{W}: Wilcoxon signed-rank test statistic. \textbf{d}: Cohen's d effect size. \textbf{Sig}: *** p<0.001, ** p<0.01, * p<0.05, ns = not significant. OFS, VeriScore, and CiteVal range 0--1 (higher is better). Linguistic Certainty ranges 1--6 (higher = more certain language). Parametric Certainty: length-normalized logprobs (closer to 0 = higher internal confidence). VeriScore is available for 3 pairs only.}
\label{tab:paired_tests_appendix}
\end{table*}

\begin{table*}[ht]
\small
\centering
\resizebox{0.9\textwidth}{!}{
\begin{tabular}{ll l rr rrr l}
\toprule
\textbf{Smaller Model} & \textbf{Larger Model} & \textbf{Metric} & \textbf{Small Med} & \textbf{Large Med}  & \textbf{W} & \textbf{p-value} & \textbf{d} & \textbf{Sig} \\
\midrule
\texttt{Llama-2-7b} & \texttt{Llama-2-70b} & OFS & 0.643 & 0.643 & 1,464,292.5 & 0.220 & -0.34 & ns \\
 &  & Ling. Cert. & 4.785 & 4.802 & 1,270,540.5 & <0.001 & -0.18 & *** \\
 &  & Par. Cert. & -0.079 & -0.096 & 577,304& <0.001 & +0.78 & *** \\
 &  & CiteVal & 0.0 & 0.133 & 498,261.5 & <0.001 & -0.21 & *** \\
\midrule
\texttt{tulu-2-7b} & \texttt{tulu-2-70b} & OFS & 0.647 & 0.682 & 1,223,338 & <0.001 & -0.13 & *** \\
 &  & Ling. Cert. & 4.812 & 4.826 & 1,324,226 & <0.001 & -0.15 & *** \\
 &  & Par. Cert. & -0.162 & -0.202 & 493,449 & <0.001 & +0.88 & *** \\
 &  & CiteVal & 0.0 & 0.333 & 96,712.5 & <0.001 & -1.06 & *** \\
\midrule
\texttt{scitulu-7b} & \texttt{scitulu-70b} & OFS & 0.651 & 0.684 & 1,296,437.5 & <0.001 & -0.10 & *** \\
 &  & Ling. Cert. & 4.854 & 4.839 & 1,383,232.5 & <0.001 & +0.11 & *** \\
 &  & Par. Cert. & -0.226 & -0.166 & 142,104 & <0.001 & -1.41 & *** \\
 &  & CiteVal & 0.0 & 0.0 & 17,692.5 & <0.001 & -1.04 & *** \\
\midrule
\texttt{Qwen2.5-7B} & \texttt{Qwen2.5-14B} & OFS & 0.684 & 0.692 & 1,332,151 & <0.001 & -0.07 & *** \\
 &  & Ling. Cert. & 4.844 & 4.859 & 1,338,484.5 & <0.001 & -0.14 & *** \\
 &  & Par. Cert. & -0.114 & -0.112 & 1,486,486 & 0.034 & -0.06 & * \\ 
 &  & CiteVal & 0.167 & 0.200 & 573,172.5 & <0.001 & -0.25 & *** \\
\midrule
\texttt{SciLitLLM1.5-7B} & \texttt{SciLitLLM1.5-14B} & OFS & 0.600 & 0.615 & 1,425,226.5 & 0.175 & -0.03 & ns \\  
 &  & Ling. Cert. & 4.806 & 4.845 & 1,226,154.5 & <0.001 & -0.21 & *** \\
 &  & Par. Cert. & -0.168 & -0.097 & 133,682 & <0.001 & -1.58 & *** \\
 &  & CiteVal & 0.0 & 0.200 & 634,676.5 & <0.001 & -0.23 & *** \\
\midrule
\texttt{Qwen3-8B} & \texttt{Qwen3-32B} & OFS & 0.706 & 0.680 & 1,340,814.5 & <0.001 & +0.05 & *** \\
 &  & Ling. Cert. & 4.893 & 4.877 & 1,290,892 & <0.001 & +0.12 & *** \\
 &  & Par. Cert. & -0.095 & -0.147 & 76,627 & <0.001 & +1.68 & *** \\
 &  & CiteVal & 0.200 & 0.333 & 655,082 & <0.001 & -0.36 & *** \\
\midrule
\texttt{S1-Base-8B} & \texttt{S1-Base-32B} & OFS & 0.630 & 0.625 & 1,505,888 & 0.748 & -0.01 & ns \\
 &  & Ling. Cert. & 4.939 & 4.898 & 764,413.5 & <0.001 & +0.51 & *** \\
 &  & Par. Cert. & -0.141 & -0.107 & 131,529 & <0.001 & -1.56 & *** \\
 &  & CiteVal & 0.118 & 0.167 & 578,765.5 & <0.001 & -0.28 & *** \\ 
\bottomrule
\end{tabular}
}
\caption{Paired Wilcoxon signed-rank tests comparing smaller to larger models within the same family. Each pair evaluated on 2,500 scientific concepts. Positive differences indicate larger model outperforms smaller model. \textbf{Small Med/Large Med} = Median scores for smaller and larger models. \textbf{W} = Wilcoxon signed-rank test statistic. \textbf{d} = Cohen's d effect size. \textbf{Sig}: *** p<0.001, ** p<0.01, * p<0.05, ns = not significant. OFS and CiteVal range 0--1 (higher is better). Linguistic Certainty ranges 1--6 (higher = more certain language). Parametric Certainty: length-normalized logprobs (closer to 0 = higher internal confidence).}
\label{tab:size_effects_appendix}
\end{table*}

\section{Detailed \texttt{VER} Human Annotation Process}
\label{app:human_annotation}

\paragraph{Topic Filters.} To ensure annotators could evaluate claims within their area of expertise, we applied domain-specific topic filters prior to entity selection. For Life Sciences, we restricted to \emph{Medicine} (250 concepts); for Engineering, \emph{Materials Science} (251 concepts); for Physical Sciences and Mathematics, \emph{Computer Science} further restricted to NLP topics (50 concepts); and for Social and Behavioural Sciences, \emph{Psychology, Sociology, or Political Science} (240 concepts). Arts and Humanities was not subject to topic filtering.

\paragraph{Concept Selection.} For each domain, a combined \texttt{VER} score per concepts was computed as the average of the OFS gamma score and VeriScore F1@K across both evaluated models. The top 5 and bottom 5 concepts per domain were selected, yielding 50 concepts in total. Combined scores for selected concepts ranged from 0.428 to 0.936 across all domains. Table~\ref{tab:human_entities} lists all selected concepts with their domain, topic filter, and combined score.

\begin{table*}[ht!]
\centering
\small
\resizebox{0.9\linewidth}{!}{
\begin{tabular}{llcc}
\toprule
\textbf{Domain} & \textbf{Scientific Concept} & \textbf{OFS} & \textbf{VeriScore}\\
\midrule
Life Sciences & $\uparrow$ \emph{the diagnosis and treatment of diabetes-related foot infections} & 0.958 & 0.915  \\
  & $\uparrow$ \emph{clinical applications of stem cell-derived exosomes} & 0.870 & 0.944 \\
 & $\uparrow$ \emph{microbiota–gut–brain axis and its therapeutic applications in neurodegenerative diseases} & 0.836 & 0.973  \\
 & $\uparrow$ \emph{extracellular vesicles for therapeutics and drug delivery} & 0.853 & 0.949  \\
 & $\uparrow$ \emph{ultra-processed foods and adverse health outcomes} & 0.871 & 0.931  \\
 & $\downarrow$ \emph{admission avoidance hospital at home} & 0.502 & 0.480  \\
 & $\downarrow$ \emph{prognostic significance of the royal marsden hospital (rmh) score in patients with cancer} & 0.624 & 0.333 \\
 & $\downarrow$ \emph{polypropylene (pp)} & 0.241 & 0.674  \\
 & $\downarrow$ \emph{pdol-based solid electrolyte in practical application} & 0.723 & 0.189  \\
 & $\downarrow$ \emph{improving patient outcomes through effective hospital administration} & 0.371 & 0.485  \\
\midrule
Arts \& Humanities & $\uparrow$ \emph{jacques derrida} & 0.909 & 0.767 \\
 & $\uparrow$ \emph{literary theory} & 0.753 & 0.904 \\
 & $\uparrow$ \emph{frantz fanon} & 0.902 & 0.749  \\
 & $\uparrow$ \emph{deconstruction} & 0.874 & 0.760 \\
 & $\uparrow$ \emph{the bhagavad gītā} & 0.736 & 0.886 \\
 & $\downarrow$ \emph{causal role theories of functional explanation} & 0.057 & 0.219  \\
 & $\downarrow$ \emph{cynosarges} & 0.193 & 0.060 \\
 & $\downarrow$ \emph{kristina wasa} & 0.090 & 0.152 \\
 & $\downarrow$ \emph{harold henry joachim} & 0.153 & 0.080  \\
 & $\downarrow$ \emph{lao sze-kwang} & 0.173 & 0.059 \\
\midrule
Phys. Sci. \& Math. & $\uparrow$ \emph{convolutional neural networks in computer vision} & 0.726 & 0.856 \\
 & $\uparrow$ \emph{autoencoders and their applications in machine learning} & 0.845 & 0.705 \\
 & $\uparrow$ \emph{deepfake detection using deep learning methods} & 0.810 & 0.696 \\
 & $\uparrow$ \emph{vision transformers and their cnn-transformer based variants} & 0.854 & 0.604 \\
 & $\uparrow$ \emph{artificial intelligence in education} & 0.647 & 0.797 \\
 & $\downarrow$ \emph{foundational models in vision} & 0.445 & 0.507 \\
 & $\downarrow$ \emph{design principles for generative ai applications} & 0.476 & 0.459  \\
 & $\downarrow$ \emph{video diffusion models} & 0.431 & 0.487  \\
 & $\downarrow$ \emph{chain of thought reasoning} & 0.435 & 0.476  \\
 & $\downarrow$ \emph{the effects of over-reliance on ai dialogue systems on students' cognitive abilities} & 0.460 & 0.400 \\
\midrule
Engineering  & $\uparrow$ \emph{flexible electronics} & 0.932 & 0.838 \\
  & $\uparrow$ \emph{industrial applications of duplex stainless steels} & 0.894 & 0.861  \\
 & $\uparrow$ \emph{silver nanoparticles} & 0.953 & 0.792  \\
& $\uparrow$ \emph{microencapsulation in food} & 0.834 & 0.875 \\
 & $\uparrow$ \emph{on-chip nanophotonics} & 0.869 & 0.782  \\
 & $\downarrow$ \emph{capacitive deionization} & 0.243 & 0.561 \\
 & $\downarrow$ \emph{biomedical high-entropy alloys} & 0.490 & 0.290 \\
& $\downarrow$ \emph{the role of carbon-based materials in enhancing the stability of perovskite solar cells} & 0.360 & 0.418  \\
 & $\downarrow$ \emph{hierarchically ordered porous materials} & 0.378 & 0.360 \\
 & $\downarrow$ \emph{alcocrfeni high-entropy alloy} & 0.425 & 0.303  \\
\midrule
Social \& Behav. Sci. & $\uparrow$ \emph{climate change and conflict} & 0.923 & 0.897  \\
 & $\uparrow$ \emph{the political power of social media platforms}  & 0.835 & 0.890  \\
 & $\uparrow$ \emph{the gig economy}  & 0.848 & 0.865  \\
  & $\uparrow$ \emph{hate speech} & 0.876 & 0.837  \\
  & $\uparrow$ \emph{systematic reviews in educational research} & 0.886 & 0.810  \\
 & $\downarrow$ \emph{meta-governance} & 0.289 & 0.487  \\
  & $\downarrow$ \emph{gender differences in attitudes to vegans/vegetarians and their food preferences [..]} & 0.429 & 0.310 \\
 & $\downarrow$ \emph{sustainable development goals deployment in business schools} & 0.292 & 0.435 \\
 & $\downarrow$ \emph{associations between critical action and positive developmental consequences [..]}  & 0.381 & 0.343  \\
 & $\downarrow$ \emph{academic approaches, methods and tools to assess circular economy at the micro level} & 0.190 & 0.481 \\
\bottomrule
\end{tabular}}
\caption{Scientific concepts selected for human evaluation per domain, with topic filter and mean OFS and VeriScore scores across all evaluated models. The top ($\uparrow$) and bottom ($\downarrow$) five concept by combined \texttt{VER} scores are chosen for annotation.}
\label{tab:human_entities}
\end{table*}

\paragraph{Claim Sampling.} For each selected concept, up to 10 atomic claims per model were sampled from the same claim sets used in the automatic evaluation pipelines. Claims from GPT-4o-mini and Sonar were interleaved in a randomised order within each concept block, and an internal identifier mapping model identity to claims was retained separately for post-hoc analysis. Model identity was not disclosed to annotators at any point during the study.

\paragraph{Annotation Interface and Guidelines.} Annotations were collected using Google sheets. Annotators were presented with one claim at a time alongside a reference document comprising the full text of the relevant survey paper and Google Search snippets retrieved by VeriScore for that claim. Each claim required two sequential decisions: first, whether the claim is scientifically check-worthy according to the provided definition; second, for check-worthy claims only, whether the claim is supported by the reference document. 

\paragraph{Annotator Profiles.} All annotators were informed of the purpose of the study and volunteered to assist without compensation. All annotators are proficient English speakers with a minimum of a bachelor's degree in their respective scientific field. In total, three annotators hold a bachelor's degree, five hold a master's degree, and two hold a doctorate. Each domain was annotated by two annotators with relevant domain expertise. Prior to annotation, all annotators received written guidelines describing the annotation tasks, label definitions, and worked examples.

\paragraph{IAA Results.} Table~\ref{tab:iaa} shows IAA using Cohen's $\kappa$ and raw agreement percentage across all five domains as well as overall. Table~\ref{tab:metric_human_agreement} shows human-metric agreement scores for both OFS and VeriScore. Table~\ref{tab:human_entities_iaa} dispays the detailed per-concept IAA result, including 10 scientific concepts from each of the five domains. We further illustrate sources of annotator (dis-)agreement on what constitutes a scientifically check-worthy claim employing the following claim samples from the annotation data. We note that claims where both annotators agreed tend to be either clearly verifiable factual statements or clearly vague and subjective: 

\paragraph{Both annotators: Scientifically Check-worthy:}
\begin{enumerate}[label=(\alph*)]
\item Micropores are less than 2 nanometers in size. (\emph{Engineering})
\item Deconstruction critiques logocentrism. (\emph{Arts and Humanities})
\item Exosomes can carry genetic material. (\emph{Life Sciences})
\item Social media platforms influence democratic processes. (\emph{Social and Behavioural Sciences})
\item Transformer architectures model long-range dependencies across multi-scale temporal dimensions. (\emph{Physical Sciences and Mathematics})
\end{enumerate}

\paragraph{Both annotators: Not Scientifically Check-worthy:}
\begin{enumerate}[label=(\alph*)]
\item Engineered materials with specific properties can meet the demands of targeted applications. (\emph{Engineering}; marked as too vague.)
\item Jacques Derrida's contributions challenge scholars to reconsider the ways in which meaning is constructed and understood. (\emph{Arts and Humanities}; marked as a subjective opinion)
\item Staff in hospitals are motivated to strive for excellence in patient care. (\emph{Life Sciences}; marked as too vague.)
\item These initiatives analyze decent work. (\emph{Social and Behavioural Sciences}; marked as too vague.)
\item CNN-transformer based models lead to more accurate representations in image classification. (\emph{Physical Sciences and Mathematics}; marked as too vague: more accurate than what?)
\end{enumerate}

On the other hand, we note that disagreements between annotators most commonly arise for claims that can be interpreted in different ways: 

\paragraph{Annotator disagreements:}
\begin{enumerate}[label=(\alph*)]
\item Silver nanoparticles provide superior antibacterial coatings for medical implants. (\emph{Engineering})
\item Cynosarges contributed to the athletic landscape of ancient Athens. (\emph{Arts and Humanities})
\item Exosomes enhance cell proliferation in regenerative medicine. (\emph{Life Sciences})
\item Climate-induced migration can create fertile ground for conflict. (\emph{Social and Behavioural Sciences}; one annotator marked as too vague while the other interpreted it as any kind of conflict.)
\item Chain of thought reasoning facilitates advancements in technology in the Physical Sciences. (\emph{Physical Sciences and Mathematics})
\end{enumerate}

We see at least three recurring sources of disagreement. First, \emph{scope ambiguity}: claims that are verifiable in a specific context but stated too broadly e.g., in Engineering, where ``superior'' is unanchored, and in Life Sciences, where the claim holds for specific conditions but is stated universally. Second, \emph{disciplinary interpretation}: claims whose check-worthiness depends on how key terms are read, e.g., in Social and Behavioural Sciences, where one annotator interpreted ``conflict'' broadly as any form of social tension while the other considered the claim too vague to verify. Third, \emph{domain boundary ambiguity}: claims that are factual but not obviously scientific in the traditional sense, e.g., in Arts and Humanities, where one annotator considered it a verifiable historical fact while the other questioned its scientific check-worthiness. These examples illustrate why defining scientific check-worthiness is non-trivial, and why IAA on this task is inherently limited without more explicit annotation guidelines per domain. They additionally show that using generic claim extraction models might not be optimal and a more in-depth study on how to define and extract scientifically check-worthy claims is needed. That being said, we acknowledge that the aforementioned explanations are only our initial remarks, and a deeper analysis on the annotators' choices might bring more insights.

\begin{table}[ht!]
\centering
\small
\begin{tabular}{lc cc cc}
\toprule
& & \multicolumn{2}{c}{\textbf{CW}} & \multicolumn{2}{c}{\textbf{F}} \\
\cmidrule(lr){3-4} \cmidrule(lr){5-6}
\textbf{Domain} & \textbf{N} & \textbf{Raw} & \textbf{$\kappa$} & \textbf{Raw} & \textbf{$\kappa$} \\
\midrule
Engineering & 200 & 0.69 & 0.31 & 0.72 & 0.39 \\
Life Sci. &  199 & 0.84 & 0.36 & 0.85 & 0.57 \\
Phy. Sci. \& Math. & 191 & 0.86 & 0.43 & 0.61 & 0.28 \\
Social \& Beh. Sci. & 187 & 0.72 & 0.24 & 0.69 & 0.19 \\
Arts \& Hum.   & 181 & 0.76 & 0.17 & 0.73 & 0.06 \\
\midrule
Overall        & 958 & 0.77 & 0.29 & 0.72 & 0.30 \\
\bottomrule
\end{tabular}
\caption{IAA for scientific check-worthiness (CW) and factuality (F) annotation, reported as raw percent agreement and Cohen's $\kappa$. N is the total number of annotated claims. Factuality agreement is computed only over claims marked check-worthy by both annotators.}
\label{tab:iaa}
\end{table}

\begin{table}[ht!]
\centering
\small
\begin{tabular}{lcccc}
\toprule
& \multicolumn{2}{c}{\textbf{OFS}} & \multicolumn{2}{c}{\textbf{VeriScore}} \\
\cmidrule(lr){2-3} \cmidrule(lr){4-5}
\textbf{Domain} & \textbf{Raw} & \textbf{$\kappa$} & \textbf{Raw} & \textbf{$\kappa$} \\
\midrule
Engineering    & 0.58 & -0.02 & 0.72 & 0.40 \\
Life Sci.     & 0.67 & 0.13 & 0.79 &  0.32  \\
Phy. Sci. \& Math. & 0.56 & 0.05 & 0.68 & 0.34 \\
Social \& Beh. Sci.   & 0.61 & 0.04 & 0.68 & 0.20 \\
Arts \& Hum. & 0.62 & 0.06 & 0.58 & 0.10 \\
\midrule
Overall & 0.61 & 0.05 & 0.69 & 0.27 \\
\bottomrule
\end{tabular}
\caption{Agreement between human factuality annotations and automatic metrics, computed over check-worthy claims only. Raw = agreement percentage; $\kappa$ = Cohen's Kappa.}
\label{tab:metric_human_agreement}
\end{table}

\begin{table*}[ht!]
\centering
\small
\resizebox{0.9\linewidth}{!}{
\begin{tabular}{llccccc}
\toprule
& & \multicolumn{2}{c}{Check-Worthiness} & \multicolumn{3}{c}{Factuality} \\
\cmidrule(lr){3-4} \cmidrule(lr){5-7} \\
\textbf{Domain} & \textbf{Scientific Concept} & \textbf{Raw} & \textbf{$\kappa$} & \textbf{N.} & \textbf{Raw} & \textbf{$\kappa$} \\
\midrule
Life Sciences & \emph{the diagnosis and treatment of diabetes-related [...]} & 0.90 & 0.74 & 14/20 & 0.93 & 0.63 \\
  & \emph{clinical applications of stem cell-derived exosomes} & 0.85 & 0.00 & 17/20 & 0.88 & 0.45 \\
 & \emph{microbiota–gut–brain axis and its therapeutic  [...]} & 0.80 & 0.27 & 15/20 & 0.87 & 0.42  \\
 & \emph{extracellular vesicles for therapeutics and drug delivery} & 0.80 & 0.23 & 15/20 & 0.73 & 0.00  \\
 & \emph{ultra-processed foods and adverse health outcomes} & 0.95 & 0.00 & 19/20 & 0.89 & 0.00 \\
 & \emph{admission avoidance hospital at home} & 1.00 & 1.00 & 15/20 & 0.80 & 0.44    \\
 & \emph{prognostic significance of the royal marsden  [...]} & 0.85 & 0.32 & 16/20 & 0.88 & 0.75 \\
 & \emph{polypropylene (pp)} & 0.70 & 0.00 & 14/20 & 0.79 & 0.53  \\
 & \emph{pdol-based solid electrolyte in practical application} & 0.70 & -0.09 & 14/20 & 0.71 & 0.39  \\
 & \emph{improving patient outcomes through effective hospital [...]} & 0.84 & 0.34 & 15/19 & 1.00 & 0.00  \\
 & \textbf{\emph{global}} & \textbf{0.84} & \textbf{0.36} & \textbf{154/199} & \textbf{0.85} & \textbf{0.57} \\ 
\midrule
Arts \& Humanities & \emph{jacques derrida} & 0.75 & 0.14 & 14/20 & 0.86 & 0.00 \\
 & \emph{literary theory} & 0.85 & -0.07 & 17/20 & 0.65 & 0.09 \\
 & \emph{frantz fanon} &  0.70 & -0.09 & 14/20 & 1.00 & 0.00  \\
 & \emph{deconstruction} & 0.75 & -0.09 & 15/20 & 0.73 & 0.00 \\
 & \emph{the bhagavad gītā} & 0.85 & 0.00 & 17/20 & 0.94 & 0.00 \\
 & \emph{causal role theories of functional explanation} & 0.60 & 0.00 & 9/15 & 0.33 & -0.23  \\
 & \emph{cynosarges} & 0.80 & 0.00 & 16/20 & 0.88 & 0.45 \\
 & \emph{kristina wasa} & 0.83 & 0.67 & 8/18 & 0.63 & -0.20 \\
 & \emph{harold henry joachim} & 0.71 & 0.00 & 10/14 & 0.50 & -0.09  \\
 & \emph{lao sze-kwang} & 0.71 & 0.00 & 10/14 & 0.40 & -0.03 \\
 & \textbf{\emph{global}} & \textbf{0.76} & \textbf{0.17} & \textbf{130/181} & \textbf{0.73} & \textbf{0.06} \\ 
\midrule
Phys. Sci. \& Math. & \emph{convolutional neural networks in computer vision} & 0.90 & 0.74 & 14/20 & 0.93 & 0.76 \\
 & \emph{autoencoders and their applications in machine learning} & 0.95 & 0.00 & 19/20 & 0.73 & 0.36 \\
 & \emph{deepfake detection using deep learning methods} & 0.60 & 0.05 & 11/20 & 0.82 & 0.61 \\
 & \emph{vision transformers and their cnn-transformer[...]} & 0.95 & 0.77 & 17/20 & 0.53 & 0.17  \\
 & \emph{artificial intelligence in education} & 0.80 & 0.00 & 16/20 & 0.63 & 0.25 \\
 & \emph{foundational models in vision} & 0.90 & 0.00 & 18/20 & 0.44 & 0.06 \\
 & \emph{design principles for generative ai applications} & 0.80 & 0.59 & 10/20 & 0.80 & 0.55  \\
 & \emph{video diffusion models} & 1.00 & 0.00 & 20/20 & 0.15 & 0.01  \\
 & \emph{chain of thought reasoning} & 0.90 & 0.00 & 18/20 & 0.61 & 0.22 \\
 & \emph{the effects of over-reliance on ai dialogue systems on [...]} & 0.82 & 0.00 & 9/11 & 0.89 & 0.77 \\
 & \textbf{\emph{global}} & \textbf{0.86} & \textbf{0.43} & \textbf{152/191} & \textbf{0.61} & \textbf{0.28} \\ 
\midrule
Engineering  & \emph{flexible electronics} & 0.70 & 0.29 & 11/20 & 0.91 & 0.62 \\
  & \emph{industrial applications of duplex stainless steels} & 0.60 & -0.05 & 11/20 & 0.64 & 0.31  \\
 & \emph{silver nanoparticles} & 0.75 & 0.50 & 7/20 & 0.71 & 0.36  \\
& \emph{microencapsulation in food} & 0.60 & 0.16 & 9/20 & 0.78 & 0.40 \\
 & \emph{on-chip nanophotonics} & 0.60 &  0.26 & 5/20 & 0.80 & 0.55 \\
 & \emph{capacitive deionization} & 0.80 & 0.55 & 12/20 & 0.75 & 0.40 \\
 & \emph{biomedical high-entropy alloys} & 0.75 & 0.50 & 9/20 & 0.44 & -0.05 \\
& \emph{the role of carbon-based materials in enhancing [...]} & 0.85 &  0.58 & 14/20 & 0.57 & 0.09 \\
 & \emph{hierarchically ordered porous materials} & 0.60 & -0.05 & 11/20 & 0.82 & 0.65 \\
 & \emph{alcocrfeni high-entropy alloy} & 0.65 & -0.17 & 13/20 & 0.77 & 0.45 \\
 & \textbf{\emph{global}} & \textbf{0.69} & \textbf{0.31} & \textbf{102/200} & \textbf{0.72} & \textbf{0.39} \\ 
\midrule
Social \& Behav. Sci. & \emph{climate change and conflict} & 0.65 & -0.21 & 13/20 & 0.77 & 0.00  \\
 & \emph{the political power of social media platforms}  & 0.80 & -0.11 & 16/20 & 0.63 & -0.20  \\
 & \emph{the gig economy}  & 0.65 & -0.09 & 13/20 & 0.62 & 0.03  \\
  & \emph{hate speech} & 0.80 & 0.23 & 15/20 & 0.93 & 0.63  \\
  & \emph{systematic reviews in educational research} & 0.65 & 0.26 & 10/20 & 0.90 & 0.62  \\
 & \emph{meta-governance} & 0.65 & 0.19 & 11/20 & 0.55 & 0.23  \\
  & \emph{gender differences in attitudes to vegans/vegetarians [...]} & 0.78  & 0.27 & 13/18 & 0.69 & 0.13 \\
 & \emph{sustainable development goals deployment in business schools} & 0.67 & 0.40 & 5/15 & 0.60 & 0.00 \\
 & \emph{associations between critical action and positive [...]}  & 0.79 & 0.53 & 8/14 & 0.25 & -0.26  \\
 & \emph{academic approaches, methods and tools to [...]} & 0.80 & 0.55 & 12/20 & 0.75 & 0.40 \\
 & \textbf{\emph{global}} & \textbf{0.72} & \textbf{0.24} & \textbf{116/187} & \textbf{0.69} & \textbf{0.19} \\ 
\bottomrule
\end{tabular}}
\caption{Inter-annotator agreement per scientific concept for check-worthiness and factuality annotation, reported as raw percent agreement and Cohen's $\kappa$. For factuality, N indicates the number of claims both annotators deemed check-worthy out of the total shown, and agreement is computed over this subset only. Domain-level global scores are computed by pooling all claims within the domain.}
\label{tab:human_entities_iaa}
\end{table*}

\section{Detailed CiteVal Human Validation Process (\texttt{ATR})}
\label{app:citeval_threshold}

\begin{figure}[ht!]
\centering
\includegraphics[width=0.6\columnwidth]{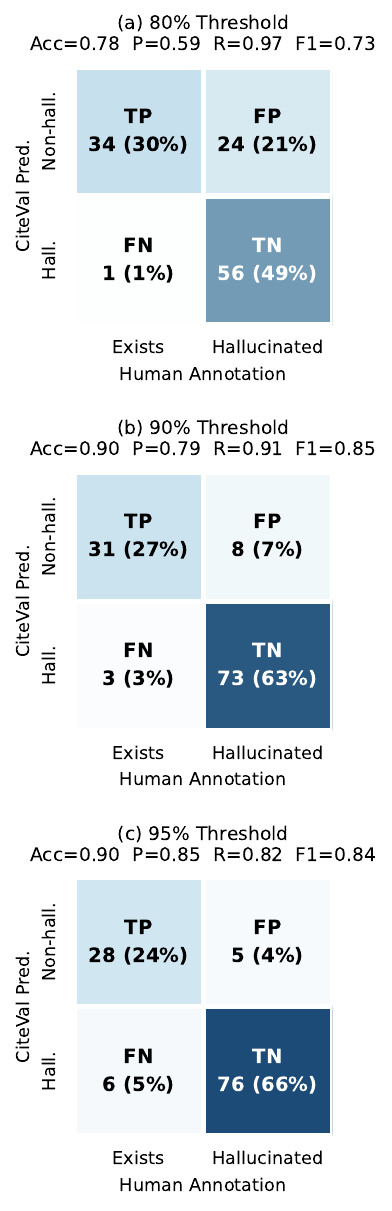}
    \caption{Human validation experiment on the same 115 extracted citations using three different threshold for title fuzzy matching: (a) 80\%; (b) 90\%; and (c) 95\%.}
    \label{fig:threshold-exp}
\end{figure}

After extracting the 115 citations and validating them, we conduct a manual audit to test three different $\alpha$ values for title fuzzy-matching when implementing CiteVal (Figure~\ref{fig:threshold-exp}). We initially tried an 80\% fuzzy-matching threshold, which was revealed to be too permissive: 21\% of citations were false positives, i.e., hallucinated citations that passed the title-matching threshold and were incorrectly marked as valid. Raising the threshold to 90\% removed the majority of false positives (24 $\rightarrow$ 8) while introducing only a small number of new false negatives (1 $\rightarrow$ 3), improving accuracy from 0.78 to 0.90 and F1 from 0.73 to 0.85. We additionally tested a 95\% threshold to assess whether further tightening would improve results. While this raised precision further and matched the 90\% threshold on overall accuracy, it did so by trading away recall (82.4\%): valid citations with minor title variation (e.g., subtitle differences or formatting artifacts) increasingly failed to match, increasing the false negative count relative to the 90\% threshold. We thus adopt $\alpha = 0.9$ as the fuzzy-matching threshold for \texttt{CiteVal} throughout this paper. However, we recommend further investigation of the optimal threshold when adapting \texttt{CiteVal}.

\end{document}